\documentclass[pdflatex,sn-vancouver-num]{sn-jnl}

\usepackage{graphicx}%
\usepackage{multirow}%
\usepackage{amsmath,amssymb,amsfonts}%
\usepackage{amsthm}%
\usepackage{mathrsfs}%
\usepackage[title]{appendix}%
\usepackage{xcolor}%
\usepackage{textcomp}%
\usepackage{manyfoot}%
\usepackage{booktabs}%
\usepackage{algorithm}%
\usepackage{algorithmicx}%
\usepackage{algpseudocode}%
\usepackage{listings}%
\usepackage{placeins}%
\usepackage{bm}%
\usepackage{subcaption}%
\usepackage{overpic}%
\usepackage{siunitx}%
\usepackage{float} %

\theoremstyle{thmstyleone}%
\theoremstyle{thmstyletwo}%

\theoremstyle{thmstylethree}%

\begin{document}

\title[Article Title]{Implementation of Quantum Implicit Neural Representation in Deterministic and Probabilistic Autoencoders for Image Reconstruction/Generation Tasks}


\author*{\fnm{Saadet Müzehher} \sur{Eren}}\email{muzehheren2000@gmail.com}

\affil{\orgdiv{Department of Physics}, \orgname{Izmir Institute of Technology}, \orgaddress{\street{Gülbahçe}, \city{Urla}, \postcode{35430}, \state{Izmir}, \country{Turkey}}}


\abstract{We propose a quantum implicit neural representation (QINR)-based autoencoder (AE) and variational autoencoder (VAE)  for image reconstruction and generation tasks. Our purpose is to demonstrate that the QINR in VAEs and AEs can transform information from the latent space into highly rich, periodic, and high-frequency features. Additionally, we aim to show that the QINR-VAE can be more stable than various quantum generative adversarial network (QGAN) models in image generation because it can address the low diversity problem.
Our quantum-classical hybrid models consist of a classical convolutional neural network (CNN) encoder and a quantum-based QINR decoder. We train the QINR-AE/VAE with binary cross-entropy with logits (BCEWithLogits) as the reconstruction loss. For the QINR-VAE, we additionally employ Kullback–Leibler divergence for latent regularization with $\beta$/capacity scheduling to prevent posterior collapse. We introduce learnable angle-scaling in data reuploading to address optimization challenges. We test our models on the MNIST, E-MNIST, and Fashion MNIST datasets to reconstruct and generate images. Our results demonstrate that the QINR structure in VAE can produce a wider variety of images with a small amount of data than various generative models that have been studied. We observe that the generated/reconstructed images from the QINR-VAE/AE are clear with sharp boundaries and details. Overall, we find that the addition of QINR-based quantum layers into the AE/VAE frameworks shows improved performance of reconstruction/generation under the constrained experimental setting relative to the specific baselines.}




\maketitle

\section{Introduction}\label{sec1}
Deep learning approaches applied to image data generally involve classification, learning meaningful representations, and generating new samples. Reconstruction-based models, such as the autoencoder (AE) \cite{michelucci2022introduction}, aim to encode the input into a compressed latent representation and then reconstruct it. Moreover, generative models such as the variational autoencoder (VAE) \cite{NEURIPS2020_e3b21256}, generative adversarial network (GAN) \cite{goodfellow2014generative}, and diffusion models \cite{10081412} aim to synthesize new patterns by learning the distribution of the training data. For instance, $\beta$-VAE \cite{higgins2017beta} is a VAE variant that scales the Kullback–Leibler (KL) regularizer by a factor $\beta$ to encourage disentangled, factorized image representations.

Although classical models still function successfully today, quantum machine learning (QML) \cite{biamonte2017quantum} algorithms have been developed to determine whether quantum computers can provide novel capabilities or efficiency for tasks related to learning. The quantum autoencoder (QAE) \cite{bravo2021quantum}, quantum variational autoencoder (QVAE), and quantum generative adversarial network (QGAN) \cite{PhysRevLett.121.040502} models were developed as quantum counterparts to classical models. 
One of these models is the parameterized quantum Wasserstein GAN (PQWGAN) \cite{10264175}. It is a hybrid quantum–classical Wasserstein GAN that uses a parameterized quantum circuit as the generator. While it focuses on generating good samples, a separate hybrid quantum–classical Wasserstein GAN, which is Quantum AnoGAN \cite{Herr_2021}, focuses on anomaly detection.
Additional studies, such as implicit neural representation (INR) \cite{dupont2021coin} and quantum implicit neural representation (QINR) \cite{pmlr-v235-zhao24}, have also been carried out to improve image processing quality. For example, Ma et al. \cite{ma2025} proposed the QINR-QGAN model with a QINR generator for image generation and suggested that the QINR approach can be considered for AEs in future work.

Instead of storing discrete samples such as pixels on a grid, the INR represents a signal as a continuous function executed by a neural network: given an input coordinate (i.e., an image's 2D position), it outputs the corresponding signal value (i.e., intensity or RGB). The INRs can model many signal types, including 1D audio/waveforms, 2D images, and 3D fields. However, for images, the important point is that the network becomes a coordinate-to-color map \cite{dupont2021coin}. The QINR has the same idea but with a parameterized quantum circuit inside the coordinate-to-pixel mapping to enhance image representation and generation. Thus, the QINR approach in the AE/VAE may become advantageous in taking a point in latent space and converting it into an instance in image space with a compact but highly expressive decoder.

Within the scope of these previous studies, we investigate whether a QINR-based data-reuploading circuit can act as a compact nonlinear spectral decoder within AE/VAE architectures. So, we introduce the QINR-VAE and QINR-AE models for image generation and reconstruction. They are quantum–classical hybrid models consisting of a classical encoder and a QINR decoder with learnable angle-scaling. The training is stabilized via binary cross-entropy with logits (BCEWithLogits) for both models, and an additional KL term for QINR-VAE with capacity/beta warm-up to improve training stability. We find that these approaches can be particularly helpful in representing the details of the images. In the QINR-QGAN model, although various input values are supplied, the GAN generator may exhibit certain difficulties in capturing the diversity of the data distribution. For example, it might become stuck during the computation and produce only a few similar samples instead of a wide variety of samples. This is known as the mode collapse problem \cite{barsha2025depth}. One of the purposes of our paper is to demonstrate that in image generation, the QINR-VAE is generally safer than the compared QGANs by showing that the QINR-VAE can mitigate the mode collapse problem.

To isolate the algorithmic behavior and facilitate controlled comparisons, we concentrate on noiseless simulations with six qubits. We qualitatively (i.e., reconstructed/generated visuals) and quantitatively (i.e., loss graphs and metrics) demonstrate the reconstruction and generation results of $28 \times 28$ pixel images from the standard MNIST \cite{6296535}, E-MNIST \cite{7966217}, and Fashion MNIST \cite{xiao2017} datasets. Since the main goal of the paper is to introduce the QINR-AE/VAE architecture rather than to perform an extensive benchmark study across many datasets, we explain the CelebA \cite{liu2015deep} experiment as supporting evidence that the proposed frameworks can be applied beyond MNIST-family datasets.
We compare the visual outputs of the QINR-VAE with those of PQWGAN \cite{10264175}, Quantum AnoGAN \cite{Herr_2021}, and the QINR-QGAN.  
We observe that the images generated by the QINR-VAE have greater intraclass diversity and better quality than the other models do. We also detect that the images reconstructed by the QINR-AE are clear. The convergence of the loss graphs shows that the optimization is stable. Additionally, we find that the results suggest that the models perform well across the considered evaluation metrics.
On the basis of these results, we conclude that our QINR-AE/VAE models with 8-D latent vectors can reconstruct/generate meaningful and distinct images with a small amount of data via class-wise training. In this sense, we use the QINR decoder as a compact spectral feature generator rather than as a replacement for optimized classical decoders.

In addition, in the Appendix, we present the results of additional experiments. We show the generated and reconstructed faces from the CelebA dataset in Appendix A. 
Here, we find that the reconstructed and generated images are regular and faded due to the limited amount of training data. In Appendix B, we compare the outputs of the reconstructed classes trained together within a single QINR-AE model, including a global angle scaling for the Fashion MNIST dataset obtained with different readouts. We observe that the multibasis readout reveals the details better than a single readout does. Finally, in Appendix C, we perform a controlled linear decoder ablation within the VAE for the MNIST dataset. Although the FID results of the classical decoder are better due to the variety, we believe that the QINR decoder generates better images in terms of visual integrity.

The outline of our paper is as follows. In Section 2, we discuss related works in this field. In Section 3, we introduce our QINR-AE and QINR-VAE models. Next, in Section 4, we present the results of our calculations. Finally, we provide a discussion in Section 5 and conclusions in Section 6.

\section{Related works}\label{sec2}
\subsection{Quantum autoencoders and quantum variational autoencoders}\label{subsec2}
The AE and VAE are latent variable models. They compress the input data into a smaller latent space, extract the important information, and return it to the original data space. In image processing, the AEs use reconstruction to learn representations. They are used in tasks such as denoising/compression with the help of reconstruction loss, particularly the mean squared error (MSE)/BCE in the pixel space. Image generation, on the other hand, is achieved by VAEs via an organized latent space with additional KL loss. There is a chance that the reconstruction/generation performed by the AE/VAE can be improved when quantum hardware is released, since quantum circuits can offer richer characteristics and probabilistic sampling. Therefore, the QAE and the QVAE were developed to address the quantum aspect.
\subsection{Quantum implicit neural representations}\label{subsubsec2}
After QINR \cite{pmlr-v235-zhao24} was developed, new models, particularly the QINR-QGAN \cite{ma2025} and the optimized quantum implicit denoising diffusion model (OQIDDM) \cite{ZHANG2025107875}, emerged. The QINR-QGAN model uses the QINR structure within the generator to generate $28\times28$ pixel images. The purpose of training with classical stabilizing methods such as the Wasserstein distance is to increase the quality of the generated images and dramatically reduce the number of trainable quantum parameters. Ma et al. \cite{ma2025} additionally considered a simplified noisy setting by applying random single-qubit rotations before the measurement. This perturbation-based approach examines the robustness to coherent deviations rather than replicating a full hardware noise process. The OQIDDM model uses the QINR in the backward denoising process. This approach produces results more rapidly when the consistency model is used. References~\cite{ma2025} and \cite{ZHANG2025107875} report the results of generated images on the MNIST, EMNIST, Fashion MNIST datasets, and in a larger-scale face generation (CelebA) scenario. Recent QINR-based neural radiance field studies show that QINR can be extended from 2D image modeling to 3D scene reconstruction and novel view synthesis \cite{cordero2025quantum}. Furthermore, quantum visual fields with neural amplitude encoding demonstrate that QINR-type representations can be extended to visual field modeling beyond standard image generation tasks \cite{wang2026quantum}.
\section{Quantum implicit neural representation of the AE/VAE}\label{sec3}
\begin{figure}[!h]
\centering
\includegraphics[width=1\linewidth]{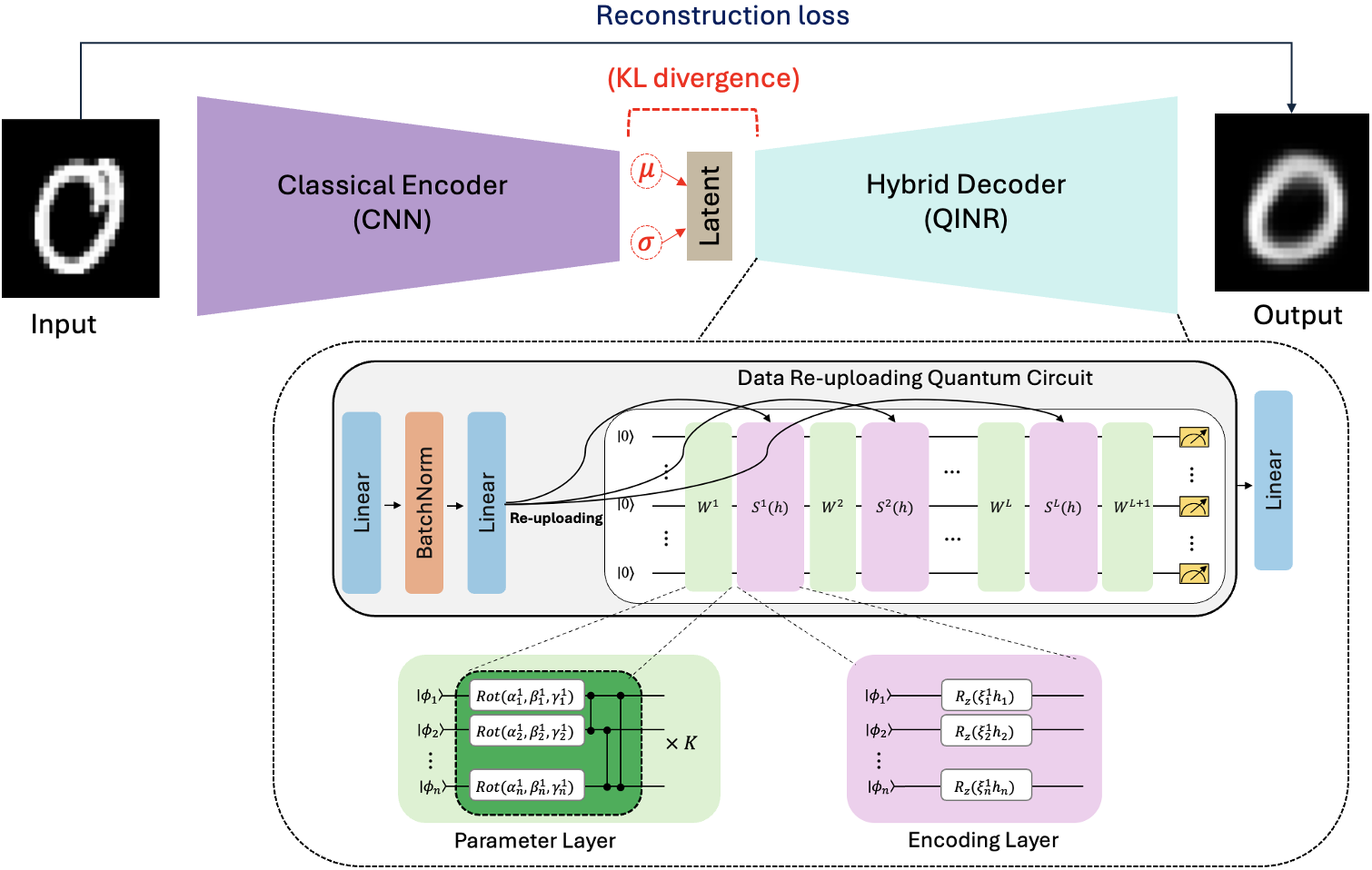}
\caption{\label{fig:Model}Schematic representation of the QINR-VAE/AE model. The QINR decoder uses a latent vector formed by the encoder to produce the output. The reconstruction loss is then calculated between the input image and the output image. In the VAE, the KL divergence is also calculated. Finally, the models are updated for optimization.}
\end{figure}

The basic design of the QINR-AE/VAE consists of two important parts: the encoder and the decoder, as shown in Fig.~\ref{fig:Model}. The classical encoder receives the input data, which in this instance is an image. As the data are compressed, their essential features are transformed into a smaller latent space by a traditional convolutional neural network (CNN) \cite{lecun1998handbook} encoder. The responsibility for allowing the conversion from the latent representation to the feature space lies with the QINR decoder, which has a hybrid structure composed of both classical and quantum layers. It contains two linear blocks with one batch normalization in between. The classical data are then passed into a quantum circuit with up to $L+1$ parameter layers and $L$ encoding layers, as shown in Fig.~\ref{fig:Model}. The parameter layers contain the Euler rotations $\mathrm{Rot}(\alpha,\beta,\gamma)$ given by
\begin{equation}
\mathrm{Rot}(\alpha,\beta,\gamma)=
\begin{pmatrix}
e^{-i(\alpha+\gamma)/2}\cos(\beta/2) & -\,e^{-i(\alpha-\gamma)/2}\sin(\beta/2)\\[4pt]
e^{i(\alpha-\gamma)/2}\sin(\beta/2) & e^{i(\alpha+\gamma)/2}\cos(\beta/2)
\end{pmatrix},
\label{eq:Rot}
\end{equation}
and controlled-Z (CZ) gates, which are 2-qubit gates denoted by a $4 \times 4$ matrix
\begin{equation}
\mathrm{CZ} \;=\;
\begin{pmatrix}
1 & 0 & 0 & 0\\
0 & 1 & 0 & 0\\
0 & 0 & 1 & 0\\
0 & 0 & 0 & -1
\end{pmatrix}
.
\label{eq:CZ}
\end{equation}
The encoding layers contain the Pauli-Z rotation $R_Z(\phi)$ gates
\begin{equation}
R_z(\phi)=
\begin{pmatrix}
e^{-i\phi/2} & 0\\
0 & e^{i\phi/2}
\end{pmatrix}
.
\label{eq:RZ}
\end{equation}
Then, the reconstruction loss is used to let the decoder produce the output correctly for both the QINR-AE and the QINR-VAE. In addition, the KL divergence is used for the QINR-VAE to organize the latent space. 

\subsection{Encoder}\label{subsec3}
The encoder is a CNN model featuring consecutive convolutional layers with batch normalization and leaky rectified linear unit (ReLU) activation functions to enhance gradient propagation. The CNN design is used as a convolutional feature extractor to compress an input image into a latent representation. A compressed abstract representation of the input image is created by progressively reducing the spatial resolution across each layer. After the final feature map is flattened, the flattened layer is passed through a fully connected layer and projected onto a deterministic latent vector $\mathbf{z}$ with dimension $d_z$. 
For the VAE, the same basic structure is used, but this time, the flattened layer is passed through two fully connected layers. One of these produces a fixed-vector mean $\bm{\mu}$, and the other produces another fixed-vector standard deviation $\bm{\sigma}$, as shown in Fig.~\ref{fig:Model}. Even with sampling, for backpropagation to work, the probabilistic $\bm{z}$ vector is obtained via the reparameterization trick, $\bm{z}= \bm{\mu}+ \bm{\sigma} \odot \bm{\epsilon}$, where $\bm{\epsilon}$ is a random noise vector drawn from the standard normal distribution.

\subsection{Decoder}\label{subsec3}
\begin{figure}[!htbp]
\centering
\includegraphics[width=1\linewidth]{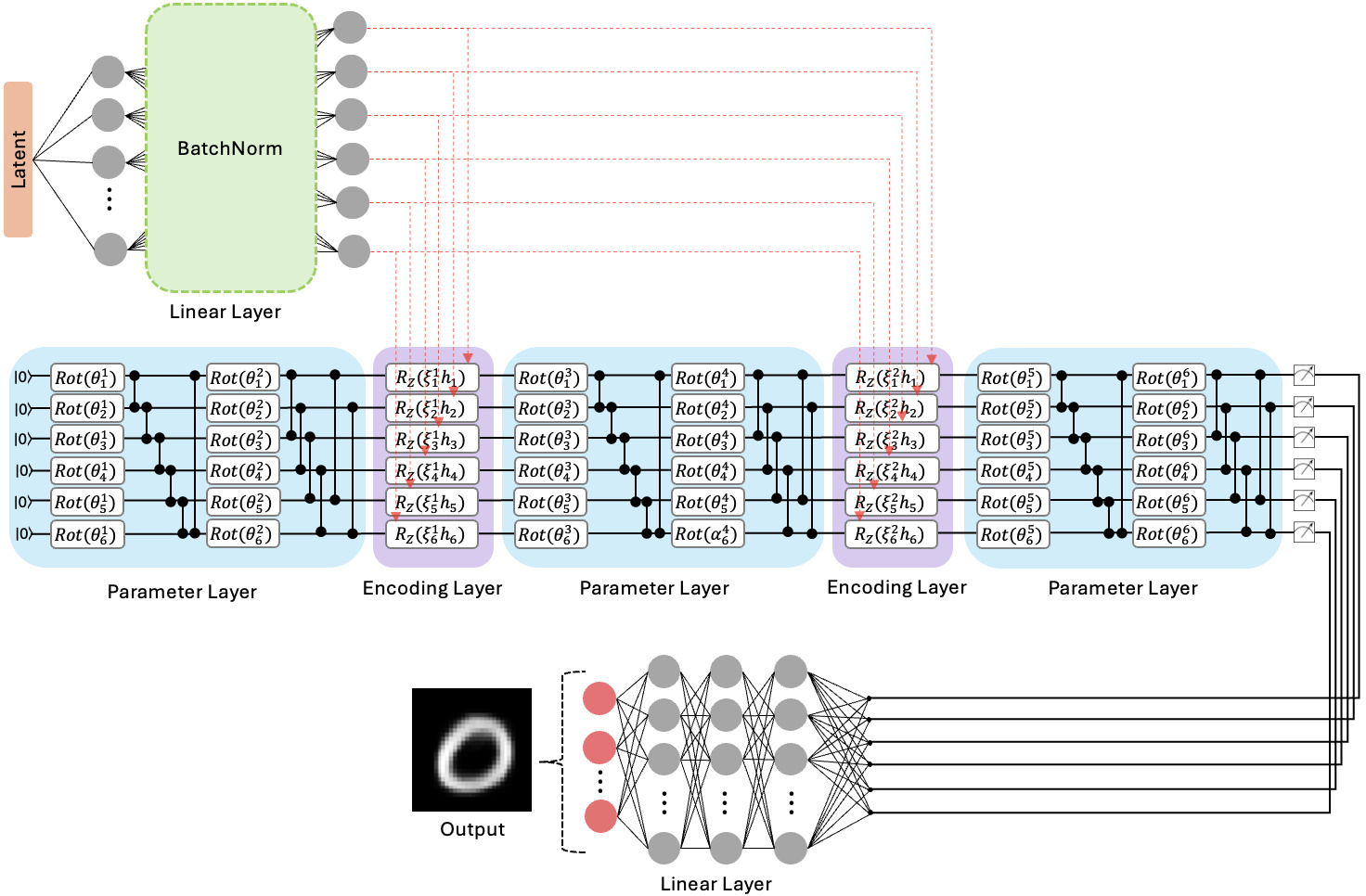}
\caption{\label{fig:Decoder}Decoder construction of the QINR–AE/VAE. It has two linear blocks with one batch normalization in between. There are a total of three parameter and two encoding layers ($L=2$). Here, $\theta_{i}^{j} \equiv (\alpha_{i}^{j}, \beta_{i}^{j}, \gamma_{i}^{j})$ is a vector representing all trainable angles for the $i$'th qubit and the $j$'th parameter layer. The data obtained through the measurement is transferred to three linear layers to produce the output.}
\end{figure}

When the latent $z_i$ reaches the hybrid layer, it first goes through batch normalization and then a higher-dimensional linear layer. In this way, we obtain
\begin{equation}
\mathbf{v}_i = BatchNorm(\mathbf{W}_1\mathbf{z}_i +\mathbf{b}_1),
\end{equation}
where $\mathbf{W_1}$ and $\mathbf{b}_1$ are the parameters of the first layer. Since quantum circuits are sensitive to angle scales, expanding the low-dimensional latent representation to a rich, high-dimensional classical representation makes the distribution of angles more stable, especially in training. A second linear angle-projection layer maps the feature vector $\mathbf{v}$ onto the learnable qubit rotation angle $\mathbf{h}$ employed in the encoding layers to match the qubit number, as shown in Fig.~\ref{fig:Decoder}. Hence, we obtain
\begin{equation}
\mathbf{h}_i = \mathbf{W}_2 \mathbf{v}_i + \mathbf{b}_2,
\end{equation}
where $\mathbf{W_2}$ and $\mathbf{b}_2$ are the parameters of the linear angle-projection layer. There are six qubits initialized in the $|0\rangle$ state. Therefore, there are six appropriate combinations for $\mathbf{h}$, which are obtained by performing learnable compression of the qubit angles. The choice of six qubits is a practical compromise between circuit expressivity and simulation cost.

\paragraph{Quantum nonlinearity (data reuploading)}Here, we use the same reuploading principle to apply the quantum block as a spectral feature extractor. Previous analyses have demonstrated that data reuploading circuits can represent Fourier series-like functions (e.g., Ref. \cite{PhysRevA.103.032430}).
A multilayer and multiqubit data reuploading circuit produces quantum features through the expectation values
\begin{equation}
f(\mathbf{h})=\langle 0|\,U^\dagger(\mathbf{h})\,O\,U(\mathbf{h})\,|0\rangle,
\label{eq:fh_expect}
\end{equation}
where the encoding layers and the alternating parameter layers form the unitary circuit given by
\begin{equation}
U(\mathbf{h})=W^{(L+1)}S^{(L)}(\mathbf{h})W^{(L)}\cdots W^{(2)}S^{(1)}(\mathbf{h})\,W^{(1)}.
\label{eq:Uh}
\end{equation}
In the implementation, $\mathbf{h}$ is used as the $R_Z$ angle from the input to the data reuploading circuit. The $\ell$'th encoding layer applies qubit-wise $R_Z$ rotations with learnable scaling $\boldsymbol{\xi}$ for six qubits:
\begin{equation}
S^\ell(\mathbf{h}) = R_Z(\xi_1^\ell h_1) \otimes R_Z(\xi_2^\ell h_2) \otimes \cdots \otimes R_Z(\xi_6^\ell h_6).
\label{eq:Sl}
\end{equation}
Especially with a few qubits, by making the reuploading scales trainable, we allow the circuit to adapt the effective input scaling across reuploads. This enhances the expressiveness and reduces manual hyperparameter tuning. The $\ell$'th parameter layer with $K=2$ can be represented as
\begin{equation}
\begin{split}
W^\ell(\bm{\theta})=
\prod_{(i,j)\in\mathcal{E}_2} \mathrm{CZ}_{ij}
\left(\mathrm{Rot}_1(\alpha_1^{2},\beta_1^{2},\gamma_1^{2}) \otimes \cdots \otimes \mathrm{Rot}_6(\alpha_6^{2},\beta_6^{2},\gamma_6^{2} \right)) \\
\quad 
\times\prod_{(i,j)\in\mathcal{E}_1} \mathrm{CZ}_{ij}
\left(\mathrm{Rot}_1(\alpha_1^{1},\beta_1^{1},\gamma_1^{1}) \otimes \cdots \otimes \mathrm{Rot}_6(\alpha_6^{1},\beta_6^{1},\gamma_6^{1}\right)).
\end{split}
\label{eq:Wl}
\end{equation}
Here, $\mathcal{E}$ denotes a pairing cluster; in other words, it shows which pairs of qubits are connected via an entangling gate.
Zhao et al. \cite{pmlr-v235-zhao24} showed that the QINR network can approximate functions in the form of a Fourier series by using this quantum circuit. The parameter $\boldsymbol{\xi}$ allows the input scaling for each qubit and reuploading step to adapt rather than being manually fixed. As a result, the effective scaled spectral terms become trainable. The scaled terms $\tilde{\boldsymbol{\lambda}}$ replace the original spectral terms $\boldsymbol{\lambda}$ as
\begin{equation} 
\tilde{\boldsymbol{\lambda}}_{j^{(\ell)}}^{(\ell)}
=
\boldsymbol{\xi}^{(\ell)} \odot \boldsymbol{\lambda}_{j^{(\ell)}}^{(\ell)}.
\end{equation}
Here, $\boldsymbol{\lambda}$ denotes the frequency/spectral components appearing in the Fourier-like expansion generated by the data-reuploading circuit. Consequently, a quantum extension of Fourier neural networks is the QINR network. Smaller values of $\boldsymbol{\xi}$ lead to smoother, lower-frequency behavior, whereas larger values allow higher-frequency variations. Finally, the output of the quantum circuit is
\begin{equation}
\mathbf{f}_i = [QC(h_i, Z_1), ... ,QC(h_i, Z_{n_q})].
\end{equation}
Even though we use the Pauli-Z in the measurement as a simple default computational readout basis, any other spin operator could have been used as a compact single readout basis. The input of the $(n+1)$'th layer will be the output $\mathbf{f}_i^{(n)}$ of the $n$'th hybrid layer so that more hybrid layers can be connected if desired. Eventually, a final linear readout is used to map $\mathbf{f}_i^{(N)}$ to the desired output space $y_i' (logits)$.

\subsection{Loss Function}\label{subsec3}
As a reconstruction loss, we use binary cross-entropy with logits (BCEWithLogitsLoss) to train the models to ensure that the decoder's output is similar to that of the target image. BCEWithLogitsLoss is given by
\begin{equation}
L_{rec}=\; -\frac{1}{N}\sum_{i=1}^{N}
\Big[\, y_i \log \sigma(y_i') \;+\; (1-y_i)\log\!\big(1-\sigma(y_i')\big) \,\Big],
\end{equation}
where $i$ is the index for all of the image elements after flattening, $N$ is the number of pixels, $y_i \in [0,1]$ is the ground-truth, and $\sigma(\cdot)$ is the sigmoid function.

For the QINR-VAE, the encoder maintains a randomly sampleable and regular latent distribution. In addition to the reconstruction loss, the KL divergence is used to regularize the encoder's posterior $q_\phi(z\mid x_i)=\mathcal{N}\!\big(\mu_i,\operatorname{diag}(\sigma_i^{2})\big)$ toward the prior $p(z)=\mathcal{N}\!\big(0, I)$. In this way, the latent space becomes smooth and can be sampled. This allows the decoder to generate new, additional valid instances instead of simply reconstructing the existing ones. The corresponding KL divergence is
\begin{equation}
L_{\mathrm{KL}}
= \frac{1}{N}\sum_{i=1}^{N}\frac{1}{2}\sum_{j=1}^{d_z}
\left(\mu_{ij}^{2}+\exp\!\left(\log \sigma_{ij}^{2}\right)-1-\log \sigma_{ij}^{2}\right),
\end{equation}
where $j$ is the latent vector index, $\sigma_{ij}$ is the standard deviation, and $\mu_{ij}$ is the mean of the encoder's latent distribution. For simpler datasets such as MNIST, the total loss for the QINR-VAE can be taken as
\begin{equation}
L = L_{\mathrm{rec}} + \beta(t) L_{\mathrm{KL}},
\end{equation}
where $\beta$ is the weight (scaling factor) of the KL divergence, which increases with time and eventually reaches one. However, for more complex datasets, the total loss includes capacity control $C(t)$, which is linearly increased to $C_{max}$, given by
\begin{equation}
L=L_{rec}+\gamma\,\left|L_{KL}-C(t)\right|,
\end{equation}
where $\gamma$ is the weight of the capacity penalty.

\subsection{Training and Optimization}\label{subsec3}
The training process of the proposed QINR-AE model for the MNIST, E-MNIST, and Fashion MNIST datasets can be explained as follows: First, the encoder $\mathrm{Enc}_{\phi}(x)$ takes the input image $x$ and compresses it into a latent vector $z$ for every mini-batch. The decoder $ \mathrm{Dec}_{\psi}(z)$ decompresses the latent to pixel-wise logits $x_{\mathrm{logits}}$. We convert the input $x$ to a range of $[-1,1]$. However, the BCE aims for the range of $[0,1]$. Therefore, the target $x_t$, which is the original image, is prepared as shown in Algorithm~\ref{alg:ae_train}. Next, the reconstruction loss is obtained as a mini-batch average by using BCEWithLogits. The gradients of the loss with respect to the encoder and decoder parameters are calculated via backpropagation for every mini-batch. To prevent gradients from becoming too large during training, global L2 norm gradient clipping is used with a threshold of $g_{max}$. Two-parameter groups are used for optimization. The separated classical layer parameters $\Theta_{\mathrm{cls}}$ are updated with the learning rate $\eta_{\mathrm{cls}}$, and the quantum layer parameters $\Theta_q$ are updated with a different learning rate $\eta_q$. In Adam's update, we maintain first- and second-moment estimates, which are controlled by the Adam hyperparameters $(\beta_1,\beta_2)$. After the parameters $\phi$ and $\psi$ are updated, the same procedure is repeated for $n_e$ epochs.
\begin{algorithm}[H]
\caption{QINR-AE Training}
\label{alg:ae_train}
\begin{algorithmic}[1]
\Require Training loader $\mathcal{L}_{\mathrm{tr}}$, epochs $n_e$, batch size $m$, grad-clip $g_{\max}$,
learning rates $\eta_{\mathrm{cls}},\eta_q$, Adam betas $(\beta_1,\beta_2)$.
\Ensure Update weights $(\phi,\psi)$ after $n_e$ epochs.
\Statex \textbf{Init:} instantiate $\mathrm{AE}=(\mathrm{Enc}_\phi,\mathrm{Dec}_\psi)$ and move to the device. Split params $\Theta_q$ (quantum) and $\Theta_{\mathrm{cls}}$ (others);
Adam with groups $(\Theta_{\mathrm{cls}},\eta_{\mathrm{cls}})$ and $(\Theta_q,\eta_q)$.
\For{$epoch=1$ to $n_e$}
  \State $\mathrm{AE}\gets \mathrm{train}()$
  \For{mini-batch $x$ from $\mathcal{L}_{\mathrm{tr}}$}
    \State $z \gets \mathrm{Enc}_{\phi}(x)$;\ \ $x_{\mathrm{logits}} \gets \mathrm{Dec}_{\psi}(z)$
    \State $x_t \gets (x+1)/2$ 
    \State $L \gets \frac{1}{m}\sum_{i=1}^{m}\mathrm{BCEWithLogits}\!\big(x_{\mathrm{logits}}^{(i)},x_t^{(i)}\big)$
    \State $\nabla \gets \nabla_{\phi,\psi} L$ ; $\nabla \gets \mathrm{ClipNorm}(\nabla,g_{\max})$
    \State $\phi,\psi \gets \mathrm{AdamStep}(\nabla)$
  \EndFor
\EndFor
\end{algorithmic}
\end{algorithm}
The logic for the algorithmic loop of the QINR-VAE is the same as that of the QINR-AE, with just a few additions. In contrast to the QINR-AE, in the QINR-VAE, the encoder generates the parameters of the latent distribution $(\mu,\log\sigma^2)$. In practice, producing the log-variance $\log\sigma^2$ is more reliable and practical than producing the standard deviation $\sigma$ directly. Therefore, $z$ can be written as $\mu + \epsilon\odot \exp(0.5\log\sigma^2)$. In addition to BCEWithLogits, KL loss is also used, as shown in Algorithm~\ref{alg:vae_train}. The variable $\beta_t$, which depends on the warm-up epoch $n_\beta$, starts small in the early epochs, and then increases to one to prevent the risk of $z$ carrying no information about $x$ anymore. This way, we ensure that the decoder does not ignore the latent; hence, we prevent posterior collapse. If capacity control is enabled, then instead of using $\beta L_{KL}$, the variable $C_t$ is pushed from zero to $C_{max}$ in $n_C$ warm-up epochs. The purpose of this is to avoid the sudden growth of KL and thus push KL gradually towards a certain information capacity over time. Apart from these, the training and optimization steps are the same as those in the QINR-VAE.
\begin{algorithm}[H]
\caption{QINR-VAE Training}
\label{alg:vae_train}
\begin{algorithmic}[1]
\Require Training loader $\mathcal{L}_{\mathrm{tr}}$, epochs $n_e$, batch size $m$, grad-clip $g_{\max}$,
learning rates $\eta_{\mathrm{cls}},\eta_q$, Adam betas $(\beta_1,\beta_2)$, capacity control $C_{\max}$, warm-up epochs $n_C$,$n_\beta$, weight $\gamma$.
\Ensure Train the QINR-VAE parameters $(\phi,\psi)$ after $n_e$ epochs.
\Statex \textbf{Init:} instantiate $\mathrm{VAE}=(\mathrm{Enc}_\phi,\mathrm{Dec}_\psi)$ and move to the device.
Split params $\Theta_q$ (quantum) and $\Theta_{\mathrm{cls}}$ (others);
Adam with groups $(\Theta_{\mathrm{cls}},\eta_{\mathrm{cls}})$ and $(\Theta_q,\eta_q)$.
\For{$epoch=1$ to $e_n$}
  \State $\mathrm{VAE}\gets \mathrm{train}()$
  \State $\beta_t \gets 1$; \quad $C_t \gets 0$
  \If{$\beta$ warm-up is enabled}
     \State $\beta_t \gets \min\!\big(1,\; epoch/n_\beta\big)$
  \EndIf
  \If{capacity control is enabled}
     \State $C_t \gets C_{\max}\cdot \min\!\big(1,\; epoch/n_C\big)$
  \EndIf
  \For{mini-batch $x$ from $\mathcal{L}_{\mathrm{tr}}$}
    \State $x_t \gets (x+1)/2$ 
    \State $(\mu,\log\sigma^2)\gets \mathrm{Enc}_{\phi}(x)$
    \State $\epsilon\sim\mathcal{N}(0,I)$;\quad $z\gets \mu + \epsilon\odot \exp(0.5\log\sigma^2)$
    \State $x_{\mathrm{logits}}\gets \mathrm{Dec}_{\psi}(z)$
    \State $L_{\mathrm{rec}} \gets \frac{1}{m}\,\mathrm{BCEWithLogits}(x_{\mathrm{logits}},x_t)$ 
    \State $L_{\mathrm{kl}} \gets \mathrm{KL}\big(q_\phi(z|x)\,\|\,\mathcal{N}(0,I)\big)$
    \If{capacity control is enabled}
        \State $L \gets L_{\mathrm{rec}} + \gamma\,\big|L_{\mathrm{kl}} - C_t\big|$
    \Else
        \State $L \gets L_{\mathrm{rec}} + \beta_t\,L_{\mathrm{kl}}$
    \EndIf
    \State $\nabla \gets \nabla_{\phi,\psi} L$;\;\; $\nabla \gets \mathrm{ClipNorm}(\nabla,g_{\max})$
    \State $(\phi,\psi)\gets \mathrm{AdamStep}(\nabla)$
  \EndFor
\EndFor
\end{algorithmic}
\end{algorithm}

\section{Experiment}\label{sec5}
The QINR-AE, QINR-VAE, and previously produced GAN \cite{ma2025} models were developed using PennyLane \cite{bergholm2022} and PyTorch \cite{NEURIPS2019}. PennyLane is one of the most widely used environments for programming quantum computers, and PyTorch is a popular deep learning framework. To obtain the quantum layers, we use PennyLane, and for the trainer, we use PyTorch. We examine the performance of QINR-VAE and QINR-AE on three datasets: MNIST, E-MNIST, and Fashion MNIST. We performed the QINR-VAE experiments on the TRUBA high-performance computing cluster using Slurm. The jobs were run using two CPU cores and 2-4 GB RAM, with a maximum wall-time limit of 12 hours. We used no GPU resources. The QINR-AE experiments were run in a Jupyter Notebook environment and were generally faster because the AE objective contains only the reconstruction loss.
Our study focuses on ideal (noiseless) simulations, where hardware is excluded. The datasets are filtered for each class, and only the first 500 samples from each class are retrieved for data preparation. We train the model using samples selected from the official training split of the dataset. Also, we perform evaluation using samples selected from the official disjoint test split. The configuration of the QINR-VAE/AE is $K=2$, $L=2$, $d_z=8$, $batch size=32$, $n_q=6$, $lr_{cls}=0.002$. We use six qubits to keep our setting directly comparable with the considered quantum generative models.
Here, $28\times28$ pixel images are generated/reconstructed by the QINR-VAE/AE (120 quantum parameters), PQWGAN (2016 quantum parameters for MNIST, E-MNIST, and 3600 quantum parameters for Fashion MNIST), Quantum AnoGAN (1008 quantum parameters), and QINR-QGAN (72 quantum parameters) models.

We use four evaluation metrics to measure the performance of the models: (1) the Fréchet inception distance (FID) \cite{heusel2017gans}, (2) the structural similarity index (SSIM) \cite {1292216}, (3) the peak signal-to-noise ratio (PSNR) \cite{5596999}, and (4) the cosine similarity \cite{zhu2018cosine}. Instead of comparing individual images, the FID metric compares the distribution of generated images with the distribution of a collection of genuine ones. Larger distances indicate a worse generative model. The SSIM metric measures the perceived value of the images by comparing the similarities in the structure of the reconstruction-type generated images to those of the source images. When its value is close to 1, the images are structurally similar. The PSNR metric measures the pixel-based difference between two images. While a larger value for the PSNR metric is typically associated with a better remake, this may not always be the case. The cosine similarity metric measures the similarity of the direction between two vectors. If it is 1, then it means that they are in the same direction.
\paragraph{QINR-VAE}We use four convolutional layers for the encoder, where each layer consists of a $3\times 3$ convolution with stride 2 and padding 1. As the calculation proceeds, the number of channels increases, and the shape chain evolves as follows: $1\times28\times28 \rightarrow32\times14\times14 \rightarrow 64\times7\times7 \rightarrow 128\times4\times4  \rightarrow 256\times  2\times 2  \rightarrow 1024$. For the decoder, we use $lr_q=0.0002$, $epoch=45$, $l \in \mathbb{R}^{128}$, and $h \in \mathbb{R}^{n_q}$ (with $n_q=6$). The results of the six measurements are fed into linear layers with 128, 512, and 784 neurons $(128 \rightarrow 512 \rightarrow 784)$. In addition, we use the following KL divergence hyperparameters. For MNIST, we use $\beta=1$  and $n_\beta = 5$. For E-MNIST, we use $C_{max}=10$, $n_C=10$, $\gamma=20$, $free bits=0.25$, and for Fashion MNIST, $C_{max}=12$, $n_C=10$,  $\gamma=10$, and $free bits=0.5$.
\paragraph{QINR-AE}The structure of the encoder is the same as that in the QINR-VAE. The only difference is that one fully connected layer maps the flattened layer to $\mathbf{z}$. For the decoder, we use the following parameters: $lr_q=0.0005$, $epoch=25$, $l \in \mathbb{R}^{256}$, $h \in \mathbb{R}^{n_q}$ (with $n_q=6$). After the measurement, the calculations proceed as follows: $256 \rightarrow 512 \rightarrow 784$. 

Beyond these basic experiments, we conducted various additional experiments, which are presented in the Appendix. In Appendix A, we show the results for the CelebA dataset. In Appendix B, we also train a model using the Fashion MNIST dataset, not in a class-wise manner but with all classes within a single model. In particular, here, we compare the effects of the measurement of (1) $\langle Z_i \rangle$ and (2) $\langle X_i \rangle$, $\langle Y_i \rangle$, $ \langle Z_i \rangle$, $\langle Z_i Z_{i+1} \rangle$ on the quality of the reconstructed images for the QINR-AE model.
\subsection{Qualitative analyses of the results}\label{subsec1}
In this section, we compare the images generated by the QINR-VAE with those obtained from the PQWGAN, Quantum AnoGAN, and QINR-QGAN models. All of the models generate recognizable characters since they are trained class-by-class (a separate model per digit or letter, etc.). Nevertheless, they display a tendency to lean towards the average because each class was trained with only 500 images. When we consider all of the datasets, we find that PQWGAN contains more background noise than the other models do. This results in more blurry images with non-uniform edges. For the Fashion MNIST dataset, we see that increasing the number of parameters reduces this nonuniformity; however, the image quality remains the same. Nevertheless, the model produces slightly different images than the QINR-QGAN and Quantum AnoGAN models do. The QINR-QGAN and Quantum AnoGAN models produce more uniform images but have very limited image diversity. Since the images produced by the two models are very close to the averages, their outputs are also very similar. On the other hand, the QINR-VAE generates cleaner, sharper, and more diverse images. We believe that it produces the most visually appealing images compared with the other models. We also find that the QINR-AE yields understandable and clear images in terms of reconstruction.

\begin{figure}[!htbp]
\centering
\includegraphics[width=1\linewidth]{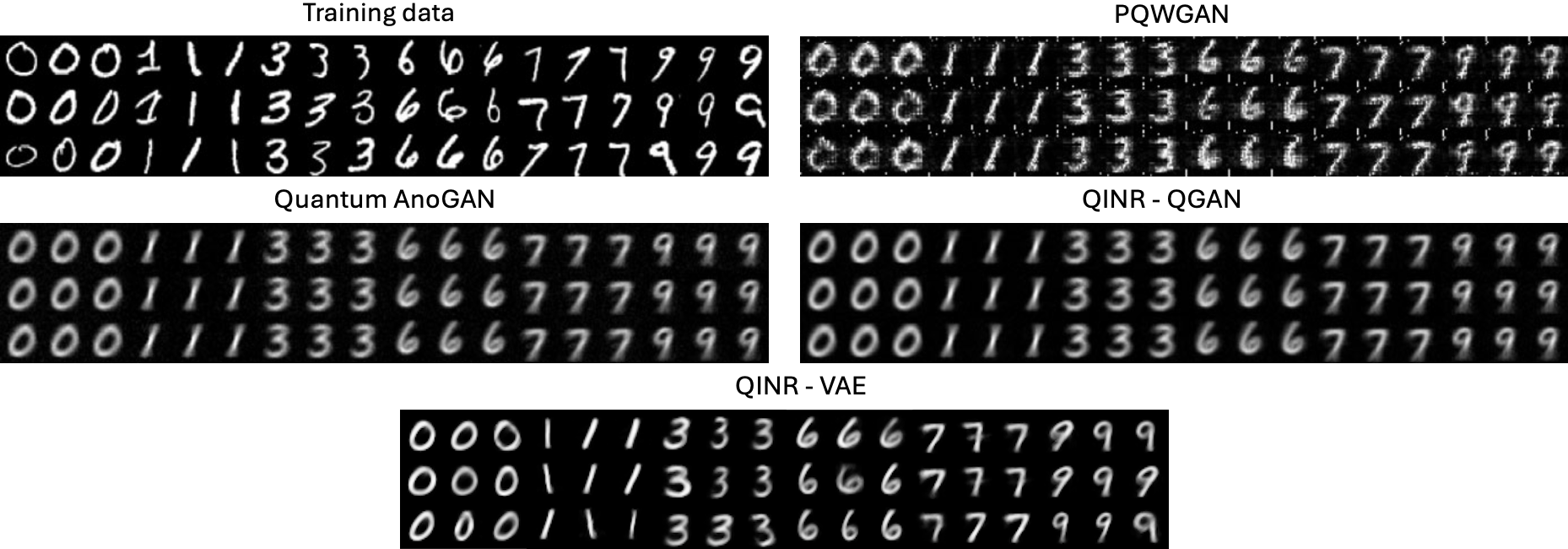}
\caption{\label{fig:MNIST}Qualitative comparison of the generative models using the MNIST dataset. Here, we observe that the Quantum AnoGAN and QINR–QGAN models generate clearer, more class-consistent results than PQWGAN does, but the results lean towards the average, indicating mode collapse. Meanwhile, the QINR-VAE generates the sharpest and most diverse samples, preserving the details.}
\end{figure}

In Fig.~\ref{fig:MNIST}, we show only the generated images of the MNIST dataset digits '0', '1', '3', '6', '7', and '9' for each model.
We observe that in the PQWGAN model, there are many irregular bright pixels and imperfections in the background. They look fuzzy, indicating poor generation quality. Tsang et al. \cite{10264175} explained this as being due to the mapping of the pixel intensities from the circuit’s final-state output distribution (amplitudes/probabilities). Consequently, the corresponding component would need to be precisely zero to produce a completely black pixel. This is challenging to accomplish with finite-depth, poorly optimized quantum circuits.
We see that although the diversity within the class is very small, it is still greater than that of the other two comparison models. The results of the Quantum AnoGAN and the QINR-QGAN are very similar. They produce cleaner and better results than PQWGAN does. However, the images are very similar within each type of digit. Both of them are visually good but do not display any variation. For the images produced by the QINR-VAE, we see that they are very clear and distinct from each other. Although the images contain small amounts of noise, they are sharper. The same digit has different writing styles, such as crossed or uncrossed '7', wider or narrower '0', slightly tilted or flat '1', etc. At first glance, we immediately see that the QINR-VAE model produces visually higher-quality images.

\FloatBarrier
\FloatBarrier

\begin{figure}[!h]
\centering
\includegraphics[width=1\linewidth]{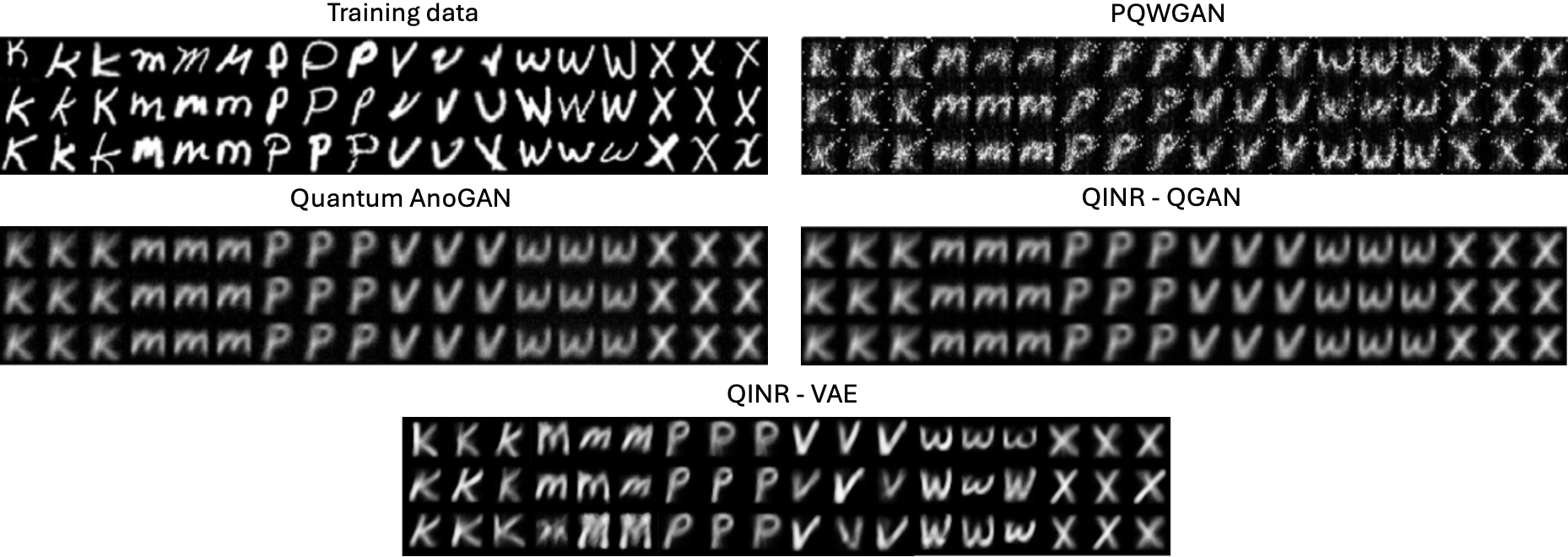}
\caption{\label{fig:E-MNIST} Same as in Figure 3, but for selected letters of the E-MNIST dataset.}
\end{figure}

Next, in Fig.~\ref{fig:E-MNIST}, we show the generated images of the E-MNIST dataset letters 'k', 'm', 'p', 'v', 'w', and 'x' for each model. We observe that in the PQWGAN model, there is again much more noise in the background, leading to fuzzy images. Quantum AnoGAN and QINR-QGAN again produce very similar results so much that their differences cannot be noticed by the naked eye. The letters in each class are quite similar, displaying mildly blurred boundaries. Both of them are visually satisfactory but do not show variance. Looking at the images produced by the QINR-VAE, we see that they are very clear and differ noticeably from each other. Although some letters, such as 'm', contain a small amount of noise, all of the generated images are sharper than those of the other models. The letters also display distinct writing styles. For example, 'w' and 'm' demonstrate better capture of both uppercase and lowercase letter diversity. 

\begin{figure}[!h]
\centering
\includegraphics[width=1\linewidth]{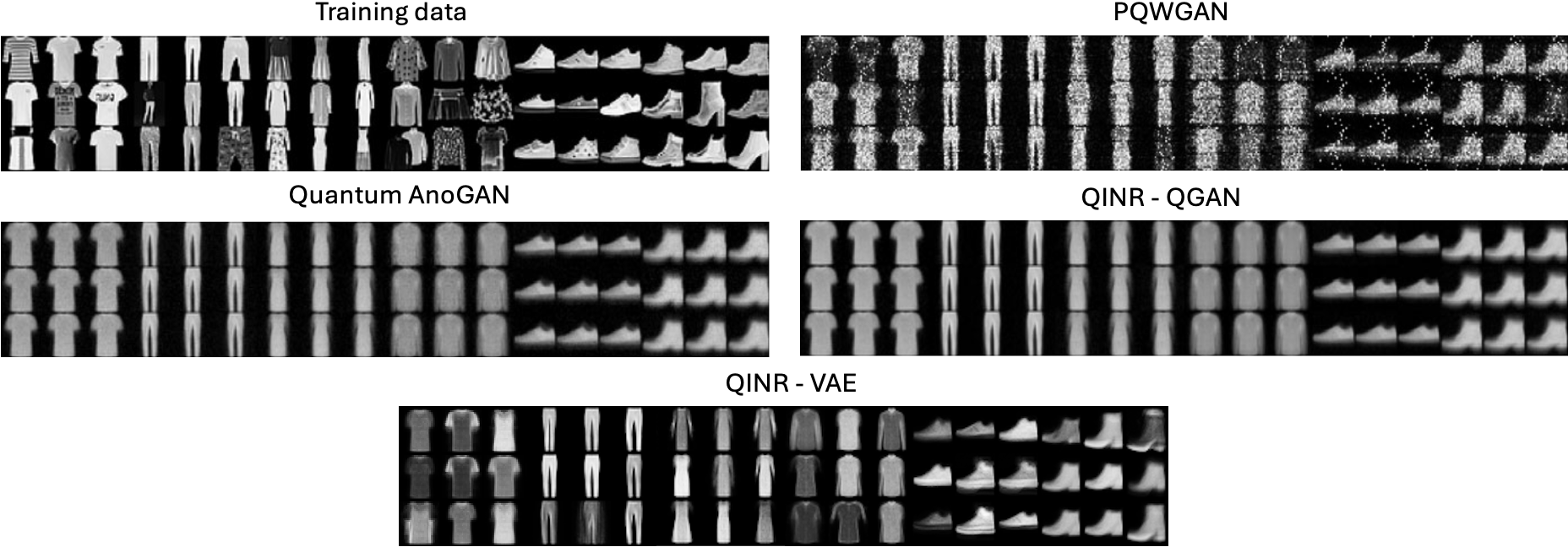}
\caption{\label{fig:F-MNIST}Same as in Figure 3, but for selected classes of the Fashion MNIST dataset.}
\end{figure}

Next, in Fig.~\ref{fig:F-MNIST}, we show the generated images of the Fashion MNIST dataset, which involves more complex image structures. In particular, we present the results for the images of 'T-shirt', 'Trouser', 'Dress', 'Shirt', 'Sneaker', and 'Ankleboot' (A.Boot). The images generated by PQWGAN exhibit some level of noise. While the background noise appears to be reduced compared with that in Figs.~\ref{fig:MNIST} and~\ref{fig:E-MNIST}, it is still noticeable even after increasing the number of quantum layers by matching the hyperparameter settings, as discussed by Tsang et al. \cite{10264175}. We see that the Quantum AnoGAN and QINR-QGAN models generate the same pattern of 'T-shirt', the same style of  'Trousers', and so on. There is also softening of the images at the edges. In contrast, we observe that the QINR-VAE model generates very sharp and distinct images within each class set. In particular, we see that it is possible to determine whether 'T-shirt' and 'Dress' have short sleeves or are sleeveless. Additionally, the color transitions for the classes are quite noticeable in grayscale. 

\begin{figure}[h!]
\centering
\includegraphics[width=0.6\linewidth,height=0.6\textheight,keepaspectratio]{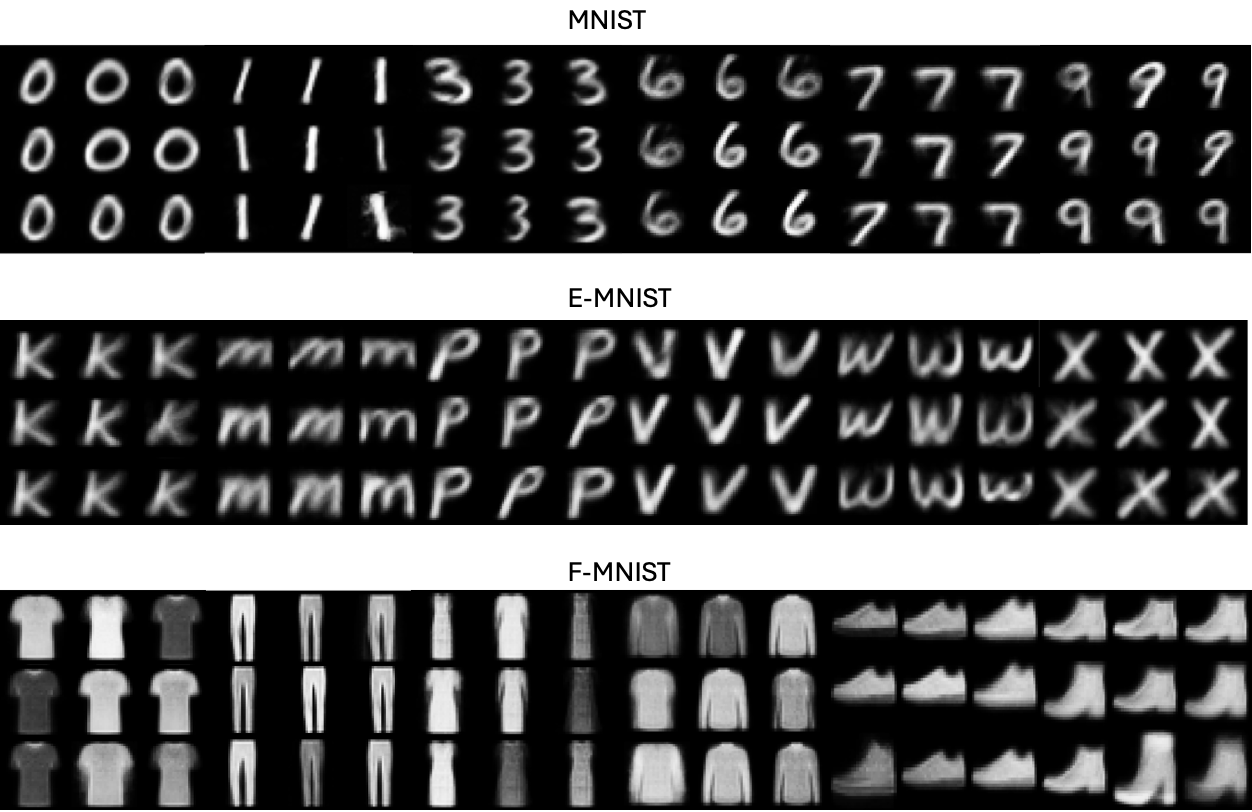}
\caption{\label{fig:AE}Reconstructed images from QINR-AE for three different datasets. We see that they are consistent with their classes and are clear with generally sharp boundaries.}
\end{figure}

Finally, Fig.~\ref{fig:AE} shows the images reconstructed by the QINR-AE model from the MNIST, E-MNIST, and Fashion MNIST datasets for the same previously selected classes. Even though increasing the number of in-class examples does not prolong the model training, we continue to use the first 500 examples as in the previous calculations. Therefore, the model results should be considered as if they were obtained with a small dataset. In the first panel of Fig.~\ref{fig:AE}, the images of the selected digits reconstructed by the QINR-AE for the MNIST dataset are shown. The examples are quite well-structured and consistent with their respective classes. Although some examples of the digit '6' are slightly blurry, the model generally succeeds in representing the corners and outlines of the digits. In the middle panel, the reconstructed images for selected letters from the E-MNIST dataset are shown. While the samples for 'k' appear more limited in terms of clarity, overall, the model is successful in clearly forming the letters. In the bottom panel, for the more complex Fashion MNIST dataset, the reconstructed clothing images are shown for the selected classes. We see that they are quite specific and detailed. In conclusion, we think that the QINR-AE model constructs images for three different datasets containing a small number of samples. It appears that the outputs of the latent representation capture important details.

\subsection{Quantitative analyses of the results}\label{subsec2}
Large-scale picture visualizations become systematic, comparable, and verifiable because of quantitative evaluations such as the loss function and metric outputs. In our paper, we calculate the losses not as an average per pixel but per image for the sum of $28 \times28 = 784$ pixels. Therefore, losses remain $\mathcal{O} (10^2)$ because of the summation over the image pixels. In this work, the FID metric is computed on ImageNet-pretrained Inception-V3 activation to maintain the FID standard specification. We also calculate the average of the SSIM, PSNR, and cosine similarity metrics. We note that the PSNR and the cosine similarity results are easily increased by the pixel average/overall similarity. The SSIM metric, however, requires structure, edge, and detail. For this reason, "memorization" can increase the values of the PSNR and the cosine similarity metrics since both fundamentally reward the pixel-level similarity.

For the QINR-VAE, we separately plot the reconstruction loss and the total (reconstruction + KL) loss to show the trade-off between the reconstruction and the regularization. We plot the total loss since the optimizer actually minimizes it when training the QINR-VAE. Additionally, we report the reconstruction term to make the optimization dynamics transparent. The results confirm that a significant part of the decrease in the total loss comes from the reconstruction. In addition, the fluctuations in the total loss for E-MNIST may originate from the capacity-controlled KL term. The mini-batch noise and the reconstruction-KL conflict may cause the KL term to fluctuate around the capacity value. Nonetheless, the amplitude of the fluctuations and the total losses decrease over time. Thus, the training can still be considered stable. For the QINR-AE, we plot only the reconstruction loss, keeping in mind that it is the only loss function. To conclude, the decreases in the total and reconstruction losses indicate that the general optimization is progressing well, and the models reconstruct the images better.

\begin{figure}[h!]
\centering
\begin{subfigure}[t]{0.33\linewidth}
  \centering
\begin{overpic}[width=\linewidth]{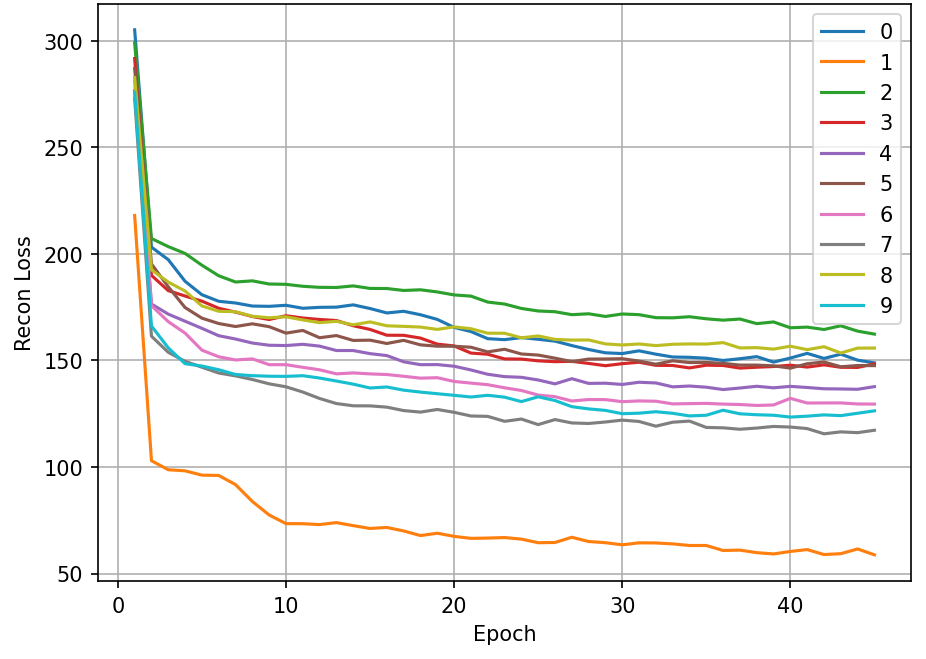}
  \put(70,55){\small\textbf{(a)}}
\end{overpic}
\phantomsubcaption\label{fig:rm}
\end{subfigure}\hfill
\begin{subfigure}[t]{0.33\linewidth}
  \centering
\begin{overpic}[width=\linewidth]{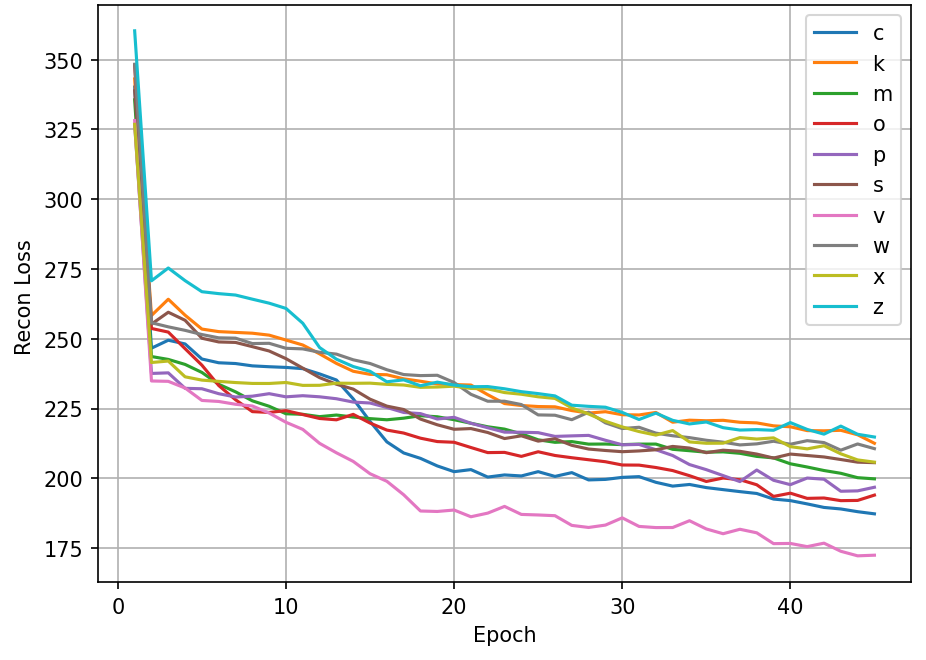}
  \put(70,55){\small\textbf{(b)}}
\end{overpic}
\phantomsubcaption\label{fig:re}
\end{subfigure}\hfill
\begin{subfigure}[t]{0.33\linewidth}
  \centering
\begin{overpic}[width=\linewidth]{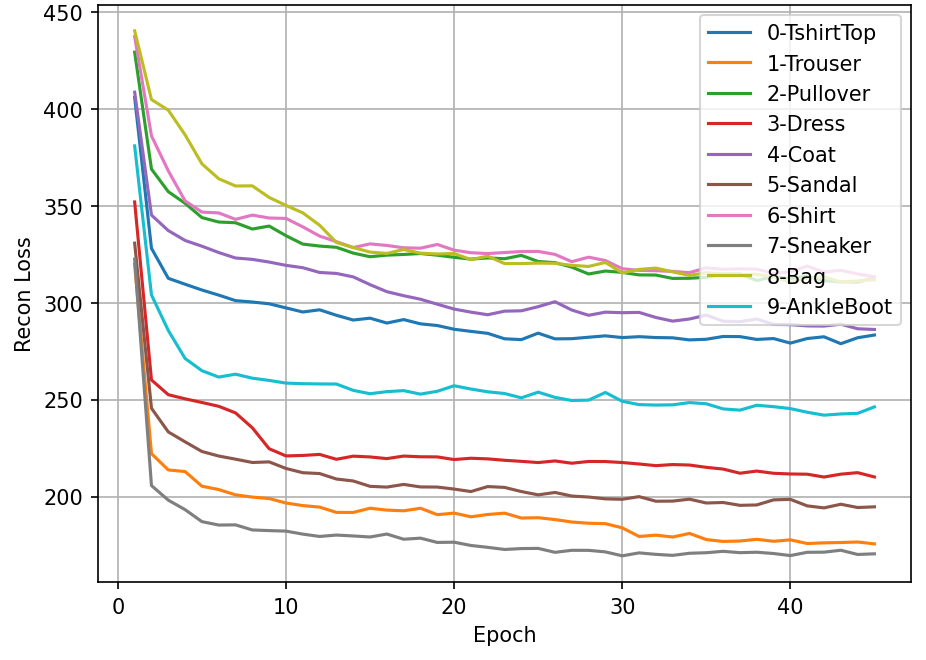}
  \put(60,55){\small\textbf{(c)}}
\end{overpic}
\phantomsubcaption\label{fig:rf}
\end{subfigure}
\caption{Reconstruction losses from the class-wise training obtained with the QINR-VAE plotted as a function of the number of epochs for the (a) MNIST (per digit), (b) E-MNIST (per letter), and (c) Fashion MNIST (per clothing) datasets. The loss trajectories indicate that the convergence is stable.}
\label{fig:7}
\end{figure}

As shown in Figures~\ref{fig:rm} and~\ref{fig:rf}, the optimization process converges when the BCEWithLogits loss steadily decreases during training and ultimately settles at the QINR-VAE model for the MNIST and Fashion MNIST datasets. In the early epochs, as the model learns the dominating structure of the data, the training loss decreases quickly. However, the pace of the decrease slows down, and the loss reaches a nearly constant value in the latter epochs. On the other hand, as shown in Fig.~\ref{fig:re}, in the later epochs for the E-MNIST dataset, the graph is generally convergent, although the loss function continues to decrease. Given the small number of trained images in the dataset, we do not increase the number of steps anymore to avoid overfitting the model.

\begin{figure}[h!]
\centering
\begin{subfigure}[t]{0.33\linewidth}
  \centering
  \begin{overpic}[width=\linewidth]{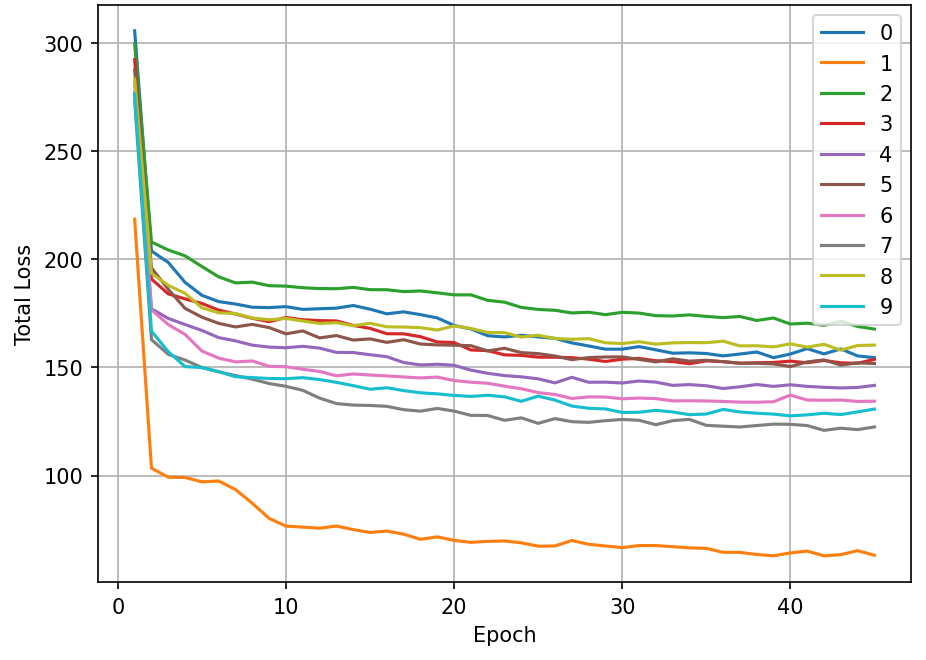}
    \put(70,55){\small\textbf{(a)}}
  \end{overpic}
  \phantomsubcaption\label{fig:tm}
\end{subfigure}\hfill
\begin{subfigure}[t]{0.33\linewidth}
  \centering
  \begin{overpic}[width=\linewidth]{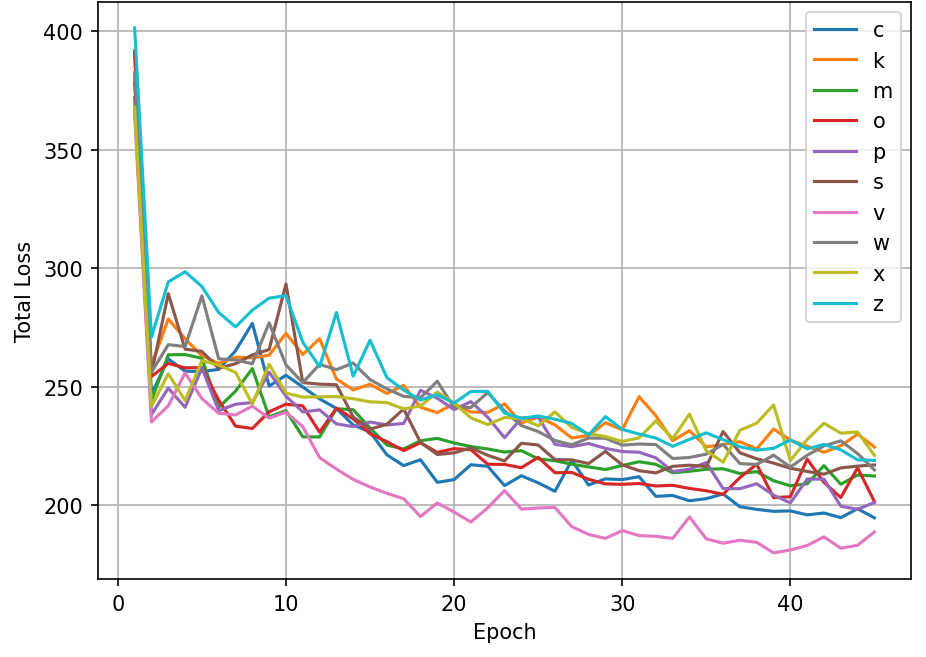}
    \put(70,55){\small\textbf{(b)}}
  \end{overpic}
  \phantomsubcaption\label{fig:te}
\end{subfigure}\hfill
\begin{subfigure}[t]{0.33\linewidth}
  \centering
  \begin{overpic}[width=\linewidth]{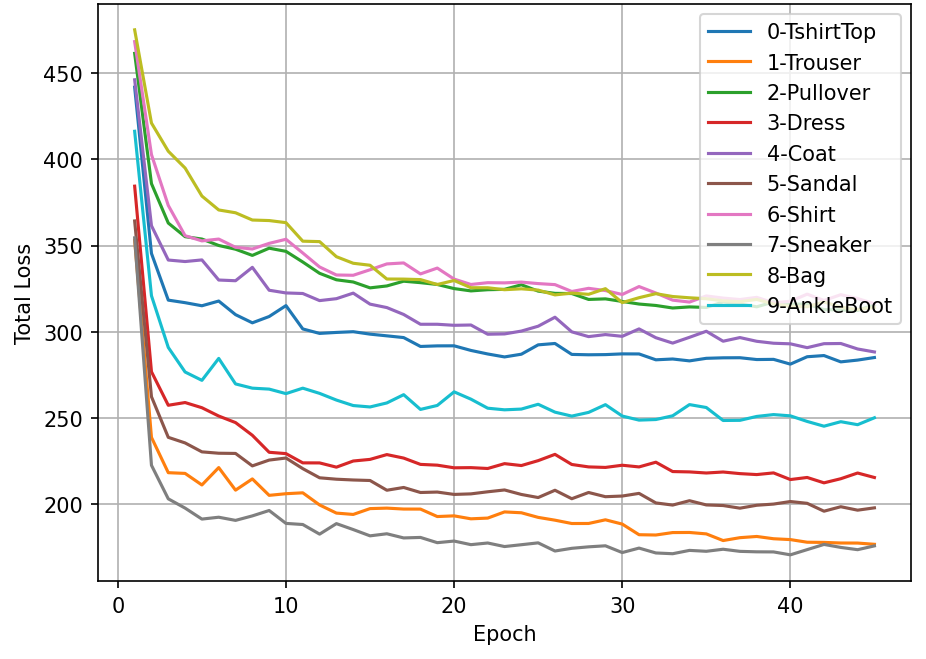}
    \put(60,55){\small\textbf{(c)}}
  \end{overpic}
  \phantomsubcaption\label{fig:tf}
\end{subfigure}
\caption{Same as in Figure 7, but for the total loss of the QINR-VAE model.}
\label{fig:8}
\end{figure}

Next, Figures~\ref{fig:tm} and~\ref{fig:tf} show that the total loss function for training decreases smoothly as the number of epochs increases for the MNIST and Fashion MNIST datasets. In Fig.~\ref{fig:te}, we observe that KL is attempting to catch up with its shifting $C_t$ target in each epoch, which causes fluctuations in the total loss for the E-MNIST dataset. Nonetheless, this does not mean that the model is not good enough. Even with the fluctuations, we consider that the model is learning quite well, since the total loss is decreasing. As the number of epochs increases, the magnitude of the fluctuations also decreases, but not monotonically. 
Considering these two types of loss graphs for the QINR-VAE model shown in Figs.~\ref{fig:7} and~\ref{fig:8}, the reduction in the reconstruction loss indicates that the model is better able to reconstruct the training data. Additionally, we see that the reduction in the total loss frequently raises the possibility that the prior samples will also improve.

\begin{figure}[h!]
\centering
\begin{subfigure}[t]{0.33\linewidth}
  \centering
\begin{overpic}[width=\linewidth]{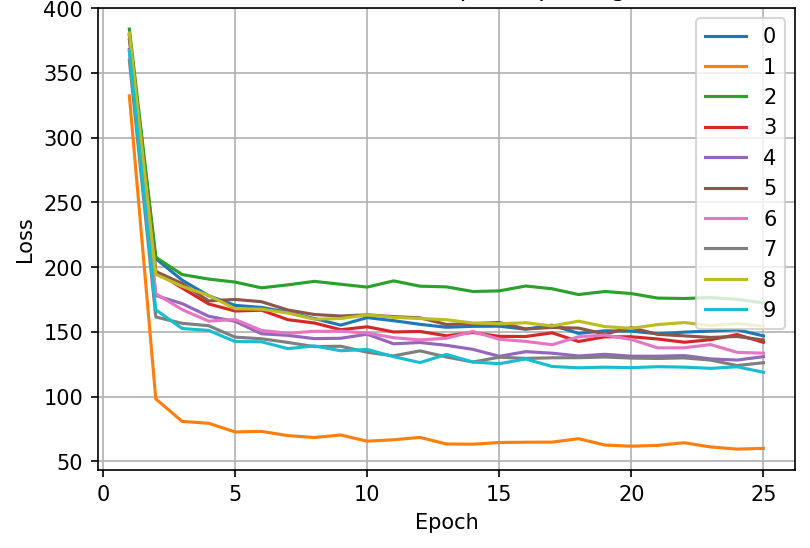}
  \put(70,55){\small\textbf{(a)}}
\end{overpic}
  \label{fig:a}
\end{subfigure}\hfill
\begin{subfigure}[t]{0.33\linewidth}
  \centering
\begin{overpic}[width=\linewidth]{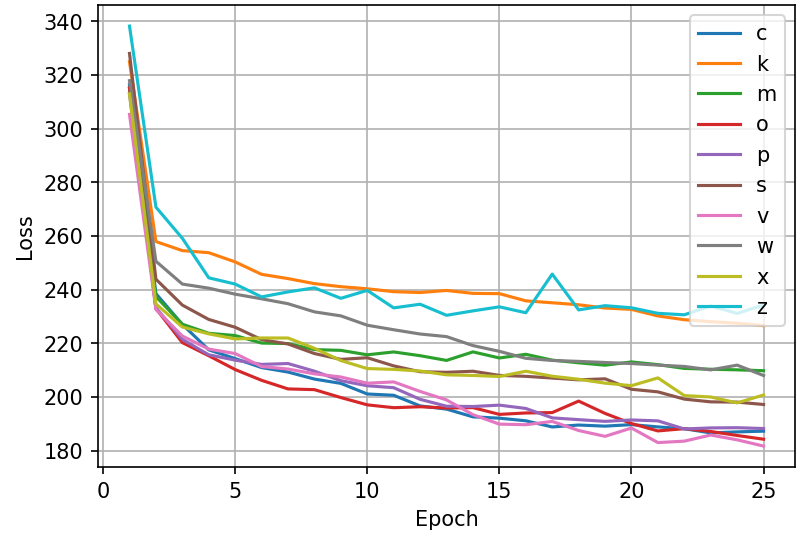}
  \put(70,55){\small\textbf{(b)}}
\end{overpic}
  \label{fig:b}
\end{subfigure}\hfill
\begin{subfigure}[t]{0.33\linewidth}
  \centering
\begin{overpic}[width=\linewidth]{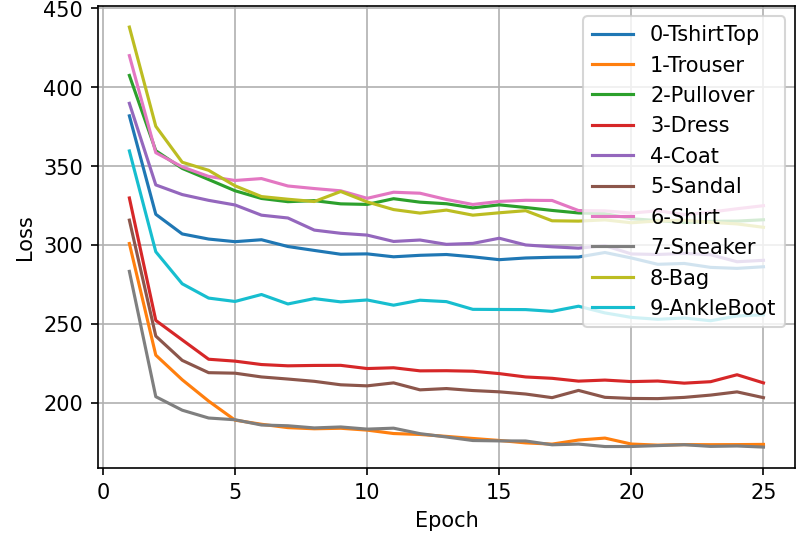}
  \put(60,55){\small\textbf{(c)}}
\end{overpic}
  \label{fig:c}
\end{subfigure}
\caption{Same as in Figure 7, but for the reconstruction loss of the QINR-AE model.}
\label{fig:9}
\end{figure}

As shown in Fig.~\ref{fig:9}, the loss for QINR-AE with BCEWithLogits has a similar performance to that of QINR-VAE, since the same reconstruction loss is used in both of these models for three different datasets. As training continues, the decrease in the loss slows down, and a plateau is reached, as expected for many ML models.

\FloatBarrier

\FloatBarrier

For prior samples of the QINR-VAE, we use the FID metric, whereas for the reconstruction samples and within the QINR-AE itself, we use the SSIM, PSNR, and cosine similarity metrics. The reasons for doing this, instead of using all of the metrics for each model, are explained below.

\paragraph{FID} Since FID is based on ImageNet-pretrained Inception features, its interpretation for small grayscale datasets should be treated with caution. Here, we use FID only as a supplementary distribution-level metric for generated samples and interpret it together with qualitative results. The FID metric implementation/protocol used in Ref. \cite{ma2025} could be further improved for clarity. Furthermore, the publicly available code employs a single-sample covariance, and the computation of a Fréchet-style distance in pixel space makes the covariance term not well-defined. This is because a proper identification of the FID metric (either inception feature or pixel space) requires multiple samples. While comparing the models, their use of the FID metric is not particularly meaningful for generative models. Moreover, a proper definition of FID is not meaningful for the QINR-AE since the primary goal of the QINR-AE model is not to generate new samples but rather to reconstruct the input. Therefore, we have made comments only about the results of the QINR-VAE and have included the remaining results as supplementary information.

Table~\ref {tab:FID-M} shows the FID metric results for the PQWGAN, Quantum AnoGAN, QINR-QGAN, and QINR-VAE models for the MNIST dataset. For the QINR-VAE, the FID takes values from 100 to 130. Table~\ref {tab:FID-E} shows the FID metric results of these models for the E-MNIST dataset. The value of the QINR-VAE ranges from 120 to 180.
Next, Table~\ref {tab:FID-F} shows the FID metric results of the same models for the Fashion MNIST dataset. For QINR-VAE, it takes values from 80 to 200.
On the basis of these tables, we draw the following conclusion: the distribution of the samples generated from the prior suggests that it deviates slightly from the true distribution of the dataset. This explains the slight blurring and the limited variety of images produced. 
\begin{table}[h!]
\centering
\setlength{\tabcolsep}{3pt}
\renewcommand{\arraystretch}{1.15}
\resizebox{0.95\linewidth}{!}{
\begin{tabular}{l*{10}{S[table-format=3.3]}}
\toprule
\textbf{Labels} &
{\textbf{0}} & {\textbf{1}} & {\textbf{2}} & {\textbf{3}} & {\textbf{4}} &
{\textbf{5}} & {\textbf{6}} & {\textbf{7}} & {\textbf{8}} & {\textbf{9}} \\
\midrule
PQWGAN          & 367.43 & 253.93 & 337.65 & 326.01 & 298.13 & 354.03 & 327.86 & 289.26 & 348.83 & 304.76 \\
Quantum AnoGAN  & 251.21 & 147.44 & 209.83 & 207.24 & 177.30 & 190.91 & 202.81 & 179.36 & 205.34 & 187.78 \\
QINR-QGAN     & 251.48 & 139.66 & 206.12 & 205.82 & 179.74 & 187.22 & 203.87 & 178.71 & 210.39 & 189.28 \\
\textbf{QINR-VAE} & 143.18 & 120.39 & 144.02 & 114.25 & 144.71 &120.42 & 120.37 &136.20 & 110.62 &122.23 \\
\bottomrule
\end{tabular}
}
\caption{FID results from different models for the MNIST dataset.}
\label{tab:FID-M}
\end{table}
\begin{table}[h!]
\centering
\setlength{\tabcolsep}{3pt}
\renewcommand{\arraystretch}{1.15}
\resizebox{0.95\linewidth}{!}{
\begin{tabular}{l*{10}{S[table-format=3.3]}}
\toprule
\textbf{Labels} &
{\textbf{C}} & {\textbf{K}} & {\textbf{M}} & {\textbf{O}} & {\textbf{P}} &
{\textbf{S}} & {\textbf{V}} & {\textbf{W}} & {\textbf{X}} & {\textbf{Z}} \\
\midrule
PQWGAN          & 392.54 & 405.99 & 373.83 & 433.54 & 385.57 & 402.32 & 409.57 & 385.43 & 382.17 & 431.91 \\
Quantum AnoGAN  & 247.72& 236.26 & 236.77 & 319.19 & 224.02 & 249.58 & 244.16 & 219.35 & 240.85 & 268.51 \\
QINR-QGAN     & 242.73 & 228.18 & 227.24 & 292.68 & 220.48 & 248.09 & 237.59 & 220.23 & 244.40 & 257.90 \\
\textbf{QINR-VAE} & 143.22 & 177.16 & 135.77 &129.68 & 125.89 & 105.92 & 130.78 & 182.00 & 137.00 & 134.08 \\
\bottomrule
\end{tabular}
}
\caption{FID results from different models for the E-MNIST dataset.}
\label{tab:FID-E}
\end{table}
\begin{table}[h!]
\centering
\setlength{\tabcolsep}{3pt}
\renewcommand{\arraystretch}{1.15}
\resizebox{0.95\linewidth}{!}{
\begin{tabular}{l*{10}{S[table-format=3.2]}}
\toprule
\textbf{Labels} &
{\textbf{T-shirt}} & {\textbf{Trouser}} & {\textbf{Pullover}} & {\textbf{Dress}} & {\textbf{Coat}} &
{\textbf{Sandal}} & {\textbf{Shirt}} & {\textbf{Sneaker}} & {\textbf{Bag}} & {\textbf{A.Boot}} \\
\midrule
PQWGAN          & 429.06 & 377.73 & 448.87 & 388.81 & 447.79 & 330.50 & 477.18 & 347.53 & 485 .90 & 425.07 \\
Quantum AnoGAN  & 268.03& 266.88 & 280.22 & 268.32 & 326.67 & 135.54 & 241.49 & 184.07 & 270.76 & 282.93 \\
QINR-QGAN     & 276.35 & 285.85 & 297.33 & 269.16 & 309.90 & 137.72 & 236.33 & 180.41 & 258.79 & 275.59 \\
\textbf{QINR-VAE} & 147.82 & 86.22 & 157.21 & 148.67 & 134.94 & 205.51 & 156.40 & 146.04 & 214.36 & 123.87 \\
\bottomrule
\end{tabular}
}
\caption{FID results from different models for the Fashion MNIST dataset.}
\label{tab:FID-F}
\end{table}
\paragraph{SSIM}Since there is no original counterpart in the ground-truth of the randomly generated sample to compare, using the SSIM metric by itself is not very meaningful in fully generative models such as the GAN. To be meaningful, it is necessary to compare two identical samples since the goal of random generation is not to produce the same image. Therefore, we comment on the SSIM metric based on only the QINR-AE and QINR-VAE (recons) reconstruction values. Nevertheless, we also show the outputs from the generative part of the QINR-VAE (prior) and the other models. In Tables~\ref{tab:SSIM-M},~\ref{tab:SSIM-E}, and~\ref{tab:SSIM-F}, we present the average SSIM metric results from the PQWGAN, Quantum AnoGAN, QINR-QGAN, QINR-VAE (reconstruction and prior), and QINR-AE for the MNIST, E-MNIST, and Fashion MNIST datasets. Although the E-MNIST success rate of the QINR-VAE/AE is slightly lower than those of the other datasets are, overall, the values are reasonable, and the core structures are preserved. However, we believe that the clarity and details can be improved further.
\begin{table}[h]
\centering
\setlength{\tabcolsep}{3pt}
\renewcommand{\arraystretch}{1.15}
\resizebox{0.95\linewidth}{!}{
\begin{tabular}{l*{10}{S[table-format=3.3]}}
\toprule
\textbf{Labels} &
{\textbf{0}} & {\textbf{1}} & {\textbf{2}} & {\textbf{3}} & {\textbf{4}} &
{\textbf{5}} & {\textbf{6}} & {\textbf{7}} & {\textbf{8}} & {\textbf{9}} \\
\midrule
PQWGAN          & 0.261 & 0.357 & 0.203 & 0.240 & 0.175 & 0.180 & 0.254 & 0.284 & 0.207 & 0.255 \\
Quantum AnoGAN  & 0.338 & 0.347 & 0.241 & 0.289 & 0.230 & 0.250 & 0.301 & 0.304 & 0.273 & 0.304 \\
QINR-QGAN     & 0.351 & 0.467 & 0.245 & 0.306 & 0.258 & 0.266 & 0.324 & 0.328 & 0.278 & 0.311 \\
QINR-VAE (prior) & 0.479 & 0.659 & 0.403& 0.459& 0.452 & 0.420 & 0.501 & 0.505& 0.485& 0.541\\
\textbf{QINR-VAE (recons)} & 0.675 & 0.860 & 0.539 & 0.637 & 0.610 &0.572 & 0.649 & 0.654 & 0.609 & 0.713 \\
\textbf{QINR-AE}  & 0.663 & 0.824 & 0.512 &0.644 & 0.644 &0.585 & 0.649 & 0.638 & 0.622 & 0.712 \\
\bottomrule
\end{tabular}
}
\caption{Average SSIM results from different models for the MNIST dataset.}
\label{tab:SSIM-M}
\end{table}
\begin{table}[h!]
\centering
\setlength{\tabcolsep}{3pt}
\renewcommand{\arraystretch}{1.15}
\resizebox{0.95\linewidth}{!}{
\begin{tabular}{l*{10}{S[table-format=3.3]}}
\toprule
\textbf{Labels} &
{\textbf{C}} & {\textbf{K}} & {\textbf{M}} & {\textbf{O}} & {\textbf{P}} &
{\textbf{S}} & {\textbf{V}} & {\textbf{W}} & {\textbf{X}} & {\textbf{Z}} \\
\midrule
PQWGAN          & 0.179 & 0.127 & 0.164 & 0.226 & 0.149 & 0.179 & 0.186 & 0.165 & 0.179 & 0.144 \\
Quantum AnoGAN  & 0.252 & 0.191 & 0.231 & 0.294 & 0.223 & 0.254 & 0.255 & 0.219 & 0.254 & 0.217 \\
QINR-QGAN     & 0.258 & 0.192 & 0.235 & 0.292 & 0.254 & 0.258 & 0.259 & 0.231 & 0.257 & 0.218 \\
QINR-VAE (prior) & 0.349 & 0.273& 0.354 & 0.443 & 0.395& 0.364& 0.380 & 0.288& 0.339& 0.268\\
\textbf{QINR-VAE (recons)} & 0.532 & 0.435 & 0.550 & 0.691 & 0.500 & 0.559 & 0.621 & 0.480 & 0.456 & 0.487 \\
\textbf{QINR-AE}  & 0.543 & 0.409 & 0.547 & 0.703 & 0.594 & 0.599 &0.545 & 0.467 & 0.495 & 0.545\\
\bottomrule
\end{tabular}
}
\caption{Average SSIM results from different models for the E-MNIST dataset.}
\label{tab:SSIM-E}
\end{table}
\begin{table}[h!]
\centering
\setlength{\tabcolsep}{3pt}
\renewcommand{\arraystretch}{1.15}
\resizebox{0.95\linewidth}{!}{
\begin{tabular}{l*{10}{S[table-format=3.2]}}
\toprule
\textbf{Labels} &
{\textbf{T-shirt}} & {\textbf{Trouser}} & {\textbf{Pullover}} & {\textbf{Dress}} & {\textbf{Coat}} &
{\textbf{Sandal}} & {\textbf{Shirt}} & {\textbf{Sneaker}} & {\textbf{Bag}} & {\textbf{A.Boot}} \\
\midrule
PQWGAN          & 0.165 & 0.303 & 0.119 & 0.180 & 0.147 & 0.095 & 0.088 & 0.205 & 0.041 & 0.143 \\
Quantum AnoGAN  & 0.223 & 0.401 & 0.177 & 0.259 & 0.225 & 0.204 & 0.143 & 0.370 & 0.088 & 0.261 \\
QINR-QGAN     & 0.230 & 0.428 & 0.180 & 0.256 & 0.225 & 0.206 & 0.142 & 0.396 & 0.090 & 0.293 \\
QINR-VAE (prior)& 0.392 & 0.631 & 0.379 & 0.477 & 0.393 & 0.345 & 0.307 & 0.557 & 0.249 & 0.416 \\
\textbf{QINR-VAE (recons)} &0.570 & 0.834 & 0.604 & 0.664 & 0.596 & 0.566 & 0.547 & 0.689 & 0.462 & 0.638 \\
\textbf{QINR-AE}  & 0.605 & 0.782 &0.582 & 0.655 & 0.603 &0.520 & 0.557 & 0.701 & 0.459 & 0.660 \\
\bottomrule
\end{tabular}
}
\caption{Average SSIM results from different models for the Fashion MNIST dataset.}
\label{tab:SSIM-F}
\end{table}

\paragraph{PSNR}As with the SSIM, in the case of the PSNR, the randomly generated images have no real target image for comparison. Therefore, we make comments on the PSNR metric based only on the QINR-AE and QINR-VAE (recons) reconstruction values. Nevertheless, the outputs from the generative part of the QINR-VAE (prior) and from the other models are shown in Tables~\ref {tab:PSNR-M},~\ref{tab:PSNR-E}, and~\ref{tab:PSNR-F}. These tables present the average PSNR metric values of the PQWGAN, Quantum AnoGAN, QINR-QGAN, QINR-VAE (reconstruction and prior), and the QINR-AE for the MNIST, E-MNIST, and Fashion MNIST datasets. Overall, we deduce that there is average pixel similarity in the QINR-VAE/AE results for different datasets. This similarity is especially enhanced for the Fashion MNIST dataset since the MSE is small. However, we observe that the results depend heavily on the task and the amount of data.
\begin{table}[h!]
\centering
\setlength{\tabcolsep}{3pt}
\renewcommand{\arraystretch}{1.15}
\resizebox{0.95\linewidth}{!}{
\begin{tabular}{l*{10}{S[table-format=3.3]}}
\toprule
\textbf{Labels} &
{\textbf{0}} & {\textbf{1}} & {\textbf{2}} & {\textbf{3}} & {\textbf{4}} &
{\textbf{5}} & {\textbf{6}} & {\textbf{7}} & {\textbf{8}} & {\textbf{9}} \\
\midrule
PQWGAN          & 10.17 & 13.29 & 10.43 & 10.88 & 11.00 & 10.27 & 11.22 & 11.60 & 10.54 & 11.42 \\
Quantum AnoGAN  & 12.11 & 15.61 & 11.88 & 12.37 & 12.79 & 12.25 & 12.90 & 13.27 & 12.48 & 13.23 \\
QINR-QGAN     & 12.14 & 15.67 & 11.91 & 12.47 & 12.88 & 12.29 & 12.90 & 13.26 & 12.41 & 13.16 \\
QINR-VAE (prior) & 10.78 & 14.05 & 11.17 & 11.68 & 11.68 & 11.18 & 11.27 &  11.83 & 11.52 & 11.73 \\
\textbf{QINR-VAE (recons)} & 13.81 & 19.05 & 13.28 & 13.91 & 14.17 & 13.52 & 13.84 & 14.72 & 13.17 & 14.57 \\
\textbf{QINR-AE}  &13.86 &18.46 & 12.99 & 14.12 & 14.57 &13.64 & 13.76 & 14.30 & 13.32 & 14.67 \\
\bottomrule
\end{tabular}
}
\caption{Average PSNR results from different models for the MNIST dataset.}
\label{tab:PSNR-M}
\end{table}
\begin{table}[h!]
\centering
\setlength{\tabcolsep}{3pt}
\renewcommand{\arraystretch}{1.15}
\resizebox{0.95\linewidth}{!}{
\begin{tabular}{l*{10}{S[table-format=3.3]}}
\toprule
\textbf{Labels} &
{\textbf{C}} & {\textbf{K}} & {\textbf{M}} & {\textbf{O}} & {\textbf{P}} &
{\textbf{S}} & {\textbf{V}} & {\textbf{W}} & {\textbf{X}} & {\textbf{Z}} \\
\midrule
PQWGAN          & 10.17 & 9.54 & 10.01 & 10.43 & 9.99 & 10.06 & 10.13 & 9.66 & 10.10 & 9.45 \\
Quantum AnoGAN  & 12.03 & 11.36 & 11.76 & 12.32 & 11.90 & 11.95 & 12.10 & 11.43 & 11.97 & 11.19 \\
QINR-QGAN     & 12.12 & 11.41 & 11.83 & 12.37 & 11.92 & 11.97 & 12.11 & 11.50 & 11.96 & 11.26 \\
QINR-VAE (prior) & 10.32 & 10.68 & 10.64 & 10.31 & 10.48 & 10.36 & 10.31 & 10.40& 10.89 & 9.77 \\
\textbf{QINR-VAE (recons)} & 13.53 & 12.45 & 12.99 & 14.25 & 12.84 &13.42 & 14.44 & 12.58 & 12.79 & 12.59 \\
\textbf{QINR-AE}  & 13.38 & 12.41 & 13.05 & 14.57 & 13.26 & 13.94 & 13.64 & 12.24 & 13.19 & 13.21 \\
\bottomrule
\end{tabular}
}
\caption{Average PSNR results from different models for the E-MNIST dataset.}
\label{tab:PSNR-E}
\end{table}
\begin{table}[h!]
\centering
\setlength{\tabcolsep}{3pt}
\renewcommand{\arraystretch}{1.15}
\resizebox{0.95\linewidth}{!}{
\begin{tabular}{l*{10}{S[table-format=3.2]}}
\toprule
\textbf{Labels} &
{\textbf{T-shirt}} & {\textbf{Trouser}} & {\textbf{Pullover}} & {\textbf{Dress}} & {\textbf{Coat}} &
{\textbf{Sandal}} & {\textbf{Shirt}} & {\textbf{Sneaker}} & {\textbf{Bag}} & {\textbf{A.Boot}} \\
\midrule
PQWGAN          & 10.86 & 12.90 & 10.19 & 11.53 & 10.71 & 10.47 & 9.72 & 11.62 & 8.90 & 10.46 \\ 
Quantum AnoGAN  & 13.33 & 15.45 & 12.56 & 13.89 & 13.17 & 13.55 & 12.33 & 15.45 & 11.19 & 13.13 \\
QINR-QGAN     & 13.36 & 15.47 & 12.49 & 13.84 & 13.17 & 13.44 & 12.34 & 15.40 & 11.22 & 13.12 \\
QINR-VAE (prior) & 10.71 & 13.25 & 10.45 & 11.62 & 10.69 & 11.90 & 10.37 & 13.21 & 9.28  &10.87 \\
\textbf{QINR-VAE (recons)} & 14.78 & 17.68 & 15.69 & 16.07 & 15.22 &14.76 & 14.94 & 16.49 & 14.31 & 15.16 \\
\textbf{QINR-AE}  & 15.65 & 15.97 & 15.29 & 15.89 & 15.60 &14.37 & 15.61 & 16.63 & 14.04 & 15.37 \\
\bottomrule
\end{tabular}
}
\caption{Average PSNR results from different models for the Fashion MNIST dataset.}
\label{tab:PSNR-F}
\end{table}

\paragraph{Cosine Similarity}As with the PSNR, the cosine similarity metric also assumes a matched target (ground-truth) in the pixel space. It would not be very meaningful to provide explanations for models that have only generative properties. Therefore, we comment on the cosine similarity metric based only on the QINR-AE and QINR-VAE (recons) reconstruction values. Tables~\ref{tab:COS-M},~\ref{tab:COS-E}, and~\ref{tab:COS-F} present the average cosine similarity metric results from PQWGAN, Quantum AnoGAN, QINR-QGAN, QINR-VAE (reconstruction and prior), and QINR-AE for the MNIST, E-MNIST, and Fashion MNIST datasets. For MNIST and E-MNIST, we observe that the main structure is preserved, but some details appear to be lost. However, for Fashion MNIST, the results are very close to the target as a pixel vector.
\begin{table}[h!]
\centering
\setlength{\tabcolsep}{3pt}
\renewcommand{\arraystretch}{1.15}
\resizebox{0.95\linewidth}{!}{
\begin{tabular}{l*{10}{S[table-format=3.3]}}
\toprule
\textbf{Labels} &
{\textbf{0}} & {\textbf{1}} & {\textbf{2}} & {\textbf{3}} & {\textbf{4}} &
{\textbf{5}} & {\textbf{6}} & {\textbf{7}} & {\textbf{8}} & {\textbf{9}} \\
\midrule
PQWGAN          & 0.894 & 0.946 & 0.890 & 0.900 & 0.905 & 0.886 & 0.908 & 0.918 & 0.892 & 0.912 \\
Quantum AnoGAN  & 0.924 & 0.968 & 0.921 & 0.929 & 0.937 & 0.928 & 0.937 & 0.943 & 0.931 & 0.942 \\
QINR-QGAN     & 0.924 & 0.986 & 0.922 & 0.931 & 0.938 & 0.928 & 0.937 & 0.943 & 0.929 & 0.941 \\
QINR-VAE (prior) & 0.660 & 0.644 & 0.622 & 0.653 & 0.614 & 0.571 & 0.632 & 0.601 & 0.677 & 0.626 \\
\textbf{QINR-VAE (recons)} & 0.826 & 0.893 & 0.767 & 0.788 & 0.783 &0.764 & 0.795 & 0.790 & 0.774 & 0.812 \\
\textbf{QINR-AE}  &0.827 & 0.877 & 0.754 & 0.798 & 0.802 &0.767 & 0.797 & 0.774& 0.779 & 0.816 \\
\bottomrule
\end{tabular}
}
\caption{Average cosine similarity results from different models for the MNIST dataset.}
\label{tab:COS-M}
\end{table}
\begin{table}[ht!]
\centering
\setlength{\tabcolsep}{3pt}
\renewcommand{\arraystretch}{1.15}
\resizebox{0.95\linewidth}{!}{
\begin{tabular}{l*{10}{S[table-format=3.3]}}
\toprule
\textbf{Labels} &
{\textbf{C}} & {\textbf{K}} & {\textbf{M}} & {\textbf{O}} & {\textbf{P}} &
{\textbf{S}} & {\textbf{V}} & {\textbf{W}} & {\textbf{X}} & {\textbf{Z}} \\
\midrule
PQWGAN          & 0.872 & 0.856 & 0.869 & 0.873 & 0.871 & 0.867 & 0.873 & 0. 859 & 0.873 & 0.847 \\
Quantum AnoGAN  & 0.915 & 0.905 & 0.912 & 0.917 & 0.916 & 0.912 & 0.919 & 0.905 & 0.916 & 0.898 \\
QINR-QGAN     & 0.917 & 0.906 & 0.913  & 0.917 & 0.916 & 0.913 & 0.919 & 0.906 & 0.916 & 0.899 \\
QINR-VAE (prior) & 0.634 & 0.642 & 0.668 & 0.723 & 0.624 & 0.659 & 0.630 & 0.629 & 0.682 & 0.610 \\
\textbf{QINR-VAE (recons)} & 0.822 & 0.767 & 0.802& 0.892& 0.780 &0.825 &0.856 & 0.776 & 0.796 & 0.799 \\
\textbf{QINR-AE}  &0.824 & 0.761 & 0.811 & 0.901 & 0.801 &0.855 & 0.825 & 0.759& 0.816 & 0.827 \\
\bottomrule
\end{tabular}
}
\caption{Average cosine similarity results from different models for the E-MNIST dataset.} 
\label{tab:COS-E}
\end{table}
\begin{table}[ht!]
\centering
\setlength{\tabcolsep}{3pt}
\renewcommand{\arraystretch}{1.15}
\resizebox{0.95\linewidth}{!}{
\begin{tabular}{l*{10}{S[table-format=3.2]}}
\toprule
\textbf{Labels} &
{\textbf{T-shirt}} & {\textbf{Trouser}} & {\textbf{Pullover}} & {\textbf{Dress}} & {\textbf{Coat}} &
{\textbf{Sandal}} & {\textbf{Shirt}} & {\textbf{Sneaker}} & {\textbf{Bag}} & {\textbf{A.Boot}} \\
\midrule
PQWGAN          & 0.848 & 0.922 & 0.803 & 0.890 & 0.838 & 0.875 & 0.786 & 0.905 & 0.754 & 0.854 \\
Quantum AnoGAN  & 0.908 & 0.954 & 0.878 & 0.935 & 0.904 & 0.931 & 0.874 & 0.957 & 0.850 & 0.916 \\
QINR-QGAN     & 0.908 & 0.953 & 0.874 & 0.934 & 0.904 & 0.930 & 0.874 & 0.956 & 0.851 & 0.917 \\
QINR-VAE (prior) & 0.856 & 0.874 & 0.879 & 0.852 & 0.884 & 0.576 & 0.847 & 0.818 & 0.790 & 0.828 \\
\textbf{QINR-VAE (recons)} & 0.916 & 0.943 & 0.931 & 0.921 & 0.937 &0.769 & 0.915 & 0.894 & 0.906 & 0.929 \\
\textbf{QINR-AE}  & 0.920 & 0.922& 0.925 &0.918 & 0.942 & 0.755 & 0.918 & 0.898 & 0.903 & 0.931 \\
\bottomrule
\end{tabular}
}
\caption{Average cosine similarity results from different models for the Fashion MNIST dataset.}
\label{tab:COS-F}
\end{table}

We find that increasing the number of epochs from 25 to 35 for the QINR-AE does not result in a significant change in the metric values. Hence, we choose the model with 25 epochs as the base model. In addition, when we increase the number of training data for each class in the Fashion MNIST dataset from 500 to 5000, we observe considerable improvements in the metrics of our models. We note that for the QINR-VAE, we are interested in improvements in the FID metric as shown in Table~\ref{tab:fivet}, and for the QINR-AE, we are interested in improvements in the SSIM, PSNR, and cosine similarity metrics as shown in Table~\ref{tab:fiveeta}.

\begin{table}[ht!]
\centering
\setlength{\tabcolsep}{3pt}
\renewcommand{\arraystretch}{1.15}
\resizebox{0.60\linewidth}{!}{
\begin{tabular}{lcccccc}
\toprule
\textbf{Labels} &
{\textbf{T-shirt}} & {\textbf{Trouser}} & {\textbf{Pullover}} & {\textbf{Dress}} & {\textbf{Coat}} &
{\textbf{Sandal}}\\
\midrule
5000 samples & 107.77& 55.43 & 110.06 & 120.09 & 91.77 & 134.91 \\
\bottomrule
\end{tabular}
}
\caption{QINR-VAE FID results for 5000 samples evaluated on the Fashion MNIST dataset.}
\label{tab:fivet}
\end{table}

\begin{table}[ht!]
\centering
\setlength{\tabcolsep}{3pt}
\renewcommand{\arraystretch}{1.15}
\resizebox{0.60\linewidth}{!}{
\begin{tabular}{lcccccc}
\toprule
\textbf{Labels} &
{\textbf{T-shirt}} & {\textbf{Trouser}} & {\textbf{Pullover}} & {\textbf{Dress}} & {\textbf{Coat}} &
{\textbf{Sandal}}\\
\midrule
PSNR  &17.18 & 20.20 & 17.11 & 17.37  & 17.45 & 15.63   \\
SSIM  & 0.694 & 0.897 & 0.696 & 0.735 & 0.702 & 0.635  \\
Cosine Similarity  & 0.939 & 0.966 & 0.947 & 0.936 & 0.958 & 0.814  \\
\bottomrule
\end{tabular}
}
\caption{QINR-AE metric results for 5000 samples evaluated on the Fashion MNIST dataset.}
\label{tab:fiveeta}
\end{table}

\section{Discussion}\label{sec5}
In this paper, we first incorporated the QINR approach into the AE and VAE architectures. Then, we analyzed the reconstruction/generation quality of the images via both qualitative and quantitative measures. We simulated the models using six qubits with 8-D latent and 120 quantum parameters. We do not claim that six qubits are strictly necessary for all possible settings. Rather, this choice is sufficient for the reconstruction and generation tasks considered in the present work while keeping the QINR decoder computationally feasible. Increasing the number of qubits may enlarge the accessible frequency spectrum and improve expressivity, but it also increases the Hilbert space dimension, simulation cost, and possible optimization difficulty. Within the models, we constructed a classical CNN encoder and a QINR decoder that contains repeated data reuploading with learnable angle-scaling and entangling layers. We used BCEWithLogits to reconstruct the images and KL divergence to regularize the latent with $\beta$/capacity scheduling. We produced $28\times28$ pixel images trained from the MNIST, E-MNIST, and Fashion MNIST datasets. We presented the reconstruction losses and the total loss for the training. We also reported the results of the SSIM, PSNR, FID, and cosine similarity metrics. We used the same class-wise training process with 500 samples per class to make comparisons with previous studies. We find that a small sample size marginally limits the performance of the models. 

On the basis of the evaluations of the images sampled from prior experiments, the experimental results show that the QINR-VAE is more successful in image generation than the QINR-QGAN, Quantum AnoGAN, and PQWGAN in terms of visual quality and diversity within the class. The generated images are clearer, more detailed, and generally more different from each other, as they should be. We also see that the performance of the QINR-AE is similar to that of the QINR-VAE in image reconstruction. Although the metric results of the QINR-VAE/AE are not perfect under the constrained experimental setting considered in this study, they are still acceptable in certain situations. The training losses decrease steadily throughout the epochs and converge to a plateau in the final phase for both of the models. This indicates that the optimization is stable and that the models reach saturation in the learning process. However, the results also show some limitations. In particular, some classes exhibit blurrier reconstructions or less diverse generated samples. These cases may arise from several factors, such as limited number of class-wise training samples. In an additional experiment, we observe an improvement in the metric values when the number of images per class is increased from 500 to 5000. Another reason can be the low-dimensional latent space and the selected parameter setting.

 Although barren plateaus \cite{larocca2025barren} are an important concern in variational quantum circuits, a full barren-plateau scaling analysis is beyond the scope of the present work. Our purpose is to evaluate a compact QINR decoder for AE/VAE-based image reconstruction/generation rather than to investigate gradient scaling with increasing qubit number or circuit depth. The decreasing loss curves, evaluation metrics, and meaningful visual outputs suggest that the deliberately chosen shallow circuit, consisting of six qubits and two data-reuploading layers, remains trainable in the considered setting. In interpreting these results, it is important to clarify that the purpose of this work is not to claim a quantum advantage over optimized classical decoders. Such a claim would be premature in the absence of reliable quantum hardware. Therefore, the QINR decoder is best viewed as a compact spectral feature generator rather than as a replacement for optimized classical decoders.

In addition to these, in the Supplementary Material, we report supplementary experiments. In Appendix A, we present the visual outputs of the CelebA dataset evaluations. Here, we point out that the QINR-AE/VAE models need to be slightly improved to increase the diversity and the sharpness of the images. Next, in Appendix B, we report the results of the extended readouts on simultaneously trained classes with global angle scaling. Here, we find that the models with extended readouts perform better than those with a single readout. Finally, to isolate the role of the QINR decoder, we include a controlled ablation against a parameter-matched linear decoder within the VAE.  We see that the QINR decoder generates more uniform but less diverse images.

\section{Conclusion}\label{sec6}
In conclusion, we found that our QINR-AE and QINR-VAE models reconstruct and generate images that are quite sharp and visually appealing under limited settings. Both of these models are good at representing details such as edges and textures. Our proposed models have demonstrated consistent performance with different datasets, and the results of the various metrics and loss terms support these findings. We also observed that the QINR-VAE generates clearer and more distinct images than other quantum GAN models do. 
Thus, we interpret this as evidence that the model did not experience mode collapse as much as it did in the other models. Even though the proposed model uses a compact quantum decoder, it exhibits a noticeable quality improvement in the considered comparisons. Therefore, we conclude that the QINR-VAE shows less evidence of mode collapse and greater image diversity than other compared QGAN models for image generation tasks. Taken together, these results demonstrate the feasibility of incorporating a QINR-based decoder into AE/VAE architectures under the considered experimental setting. Finally, our results support the possibility of developing more competitive innovations in image reconstruction/generation within the scope of QML.

In the future, further studies can be carried out to enhance the models, such as by improving the metric values, visual quality, and diversity of the images. We note that our evaluations focus on training stability and qualitative visual outputs. Broader evaluation on more diverse datasets, a systematic comparison against optimized convolutional decoders, barren plateau scaling analysis with larger qubit numbers and deeper circuits, and the robustness of the results under realistic hardware noise models or execution on quantum devices are left for future research.

\bmhead{Acknowledgements}The author gratefully thanks Yalın Baştanlar, Nejat Bulut, Ruha Uğraş Erdoğan, and Ferit Acar Savacı for their helpful discussions and comments. The numerical calculations reported in this paper were partially performed at TUBITAK ULAKBIM, High Performance and Grid Computing Center (TRUBA resources)
\bmhead{Funding}This research received no funding.
\bmhead{Conflict of interest}The author declares no competing interests.
\bmhead{Ethics approval and consent to participate}Not applicable.

\bmhead{Data availability}The data that support the findings of this study are openly available at the following URLs/DOI: MNIST (\url{https://doi.org/10.24432/C53K8Q}); E-MNIST (\url{https://www.nist.gov/itl/products-and-services/emnist-dataset}); Fashion-MNIST (\url{https://github.com/zalandoresearch/fashion-mnist}); and CelebA (\url{https://mmlab.ie.cuhk.edu.hk/projects/CelebA.html}). All other data that support the findings of this study are included within the article and any supplementary information files.
\bmhead{Author contributions}The author was solely responsible for the conception, design, data collection, analysis, and writing of the manuscript.

\begin{appendices}

\section{QINR-AE/VAE results on the CelebA dataset}\label{secA1}
In Appendix A, we examine the performance of the QINR-AE and QINR-VAE models on the CelebA dataset as a non-MNIST family dataset.
Our goal here is to reconstruct/generate $78 \times 64$ pixel images via the QINR-AE/VAE. In particular, we select the top 10 individuals with the most examples in CelebA. Since the number of training sets is small, in addition to the first 20 original images, we include a total of 500 images in the training set with 480 tensor-level augmentations. In the implementation, we use $L=3$ and $K=3$. 
The higher-dimensional space has 512 neurons. After the measurements, the calculations proceed as $512 \rightarrow 128 \rightarrow 512 \rightarrow4992$. In addition, we use MSE as the reconstruction loss. For the QINR-VAE, we choose the parameters $C_{max}=30$, $n_C=10$, and $\gamma=10$ with $free bits=0.25$, and $epochs=20$. 
\FloatBarrier
\begin{figure}[!h]
\centering
\includegraphics[width=0.9\linewidth]{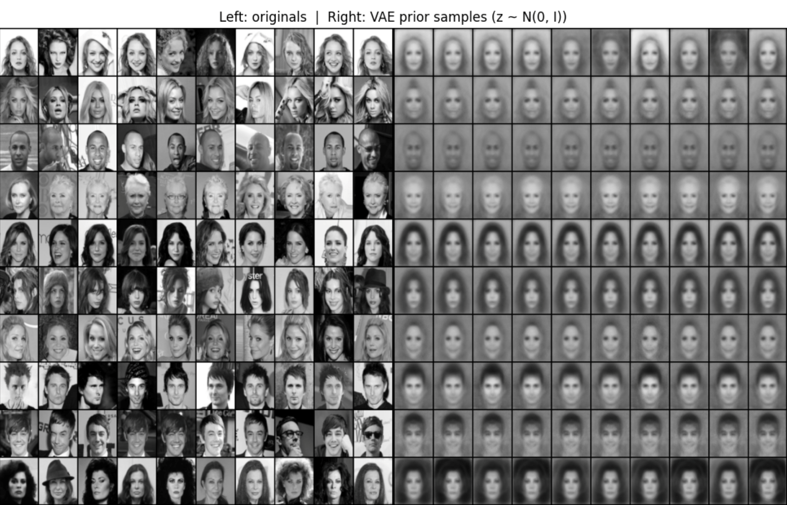}
\caption{\label{fig:CelebVAE}CelebA faces of the top 10 individuals generated by QINR–VAE with $78 \times 64$ pixels.}
\end{figure}
\begin{figure}[!h]
\centering
\includegraphics[width=0.9\linewidth]{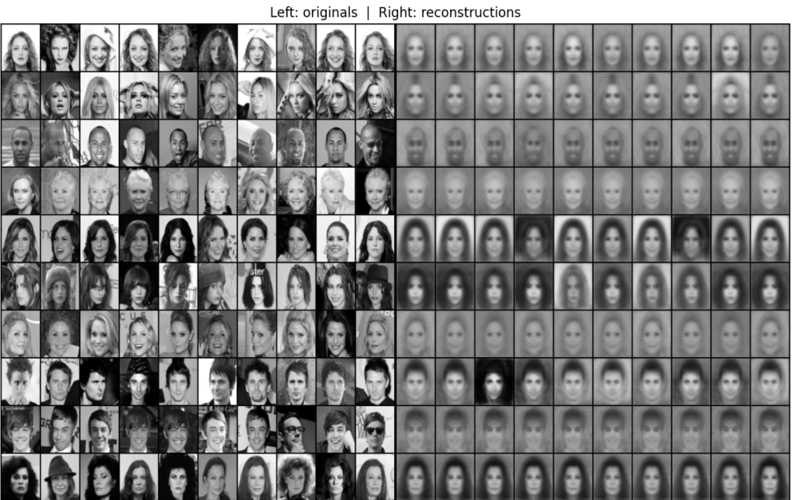}
\caption{\label{fig:CelebAE}CelebA faces of the top 10 individuals reconstructed by QINR-AE with $78 \times 64$ pixels.}
\end{figure}
\FloatBarrier

The faces generated by the QINR-VAE are shown in Fig.~\ref{fig:CelebVAE}, whereas the results of the QINR-AE are shown in  Fig.~\ref{fig:CelebAE}. Since data augmentation cannot significantly increase the diversity of the images, we expect that the reconstructed/generated images will be uniform and vague. 
We see that this is indeed the case for the images shown in Figs.~\ref{fig:CelebVAE} and~\ref{fig:CelebAE}. 
Here, we also observe that the images reconstructed by the QINR-AE are sharper than those generated by the QINR-VAE.

\section{Simultaneously trained classes with different types of readouts }\label{secA2}
\begin{figure}[h]
    \centering
    \begin{subfigure}[b]{0.48\textwidth}
        \centering
        \includegraphics[width=\linewidth]{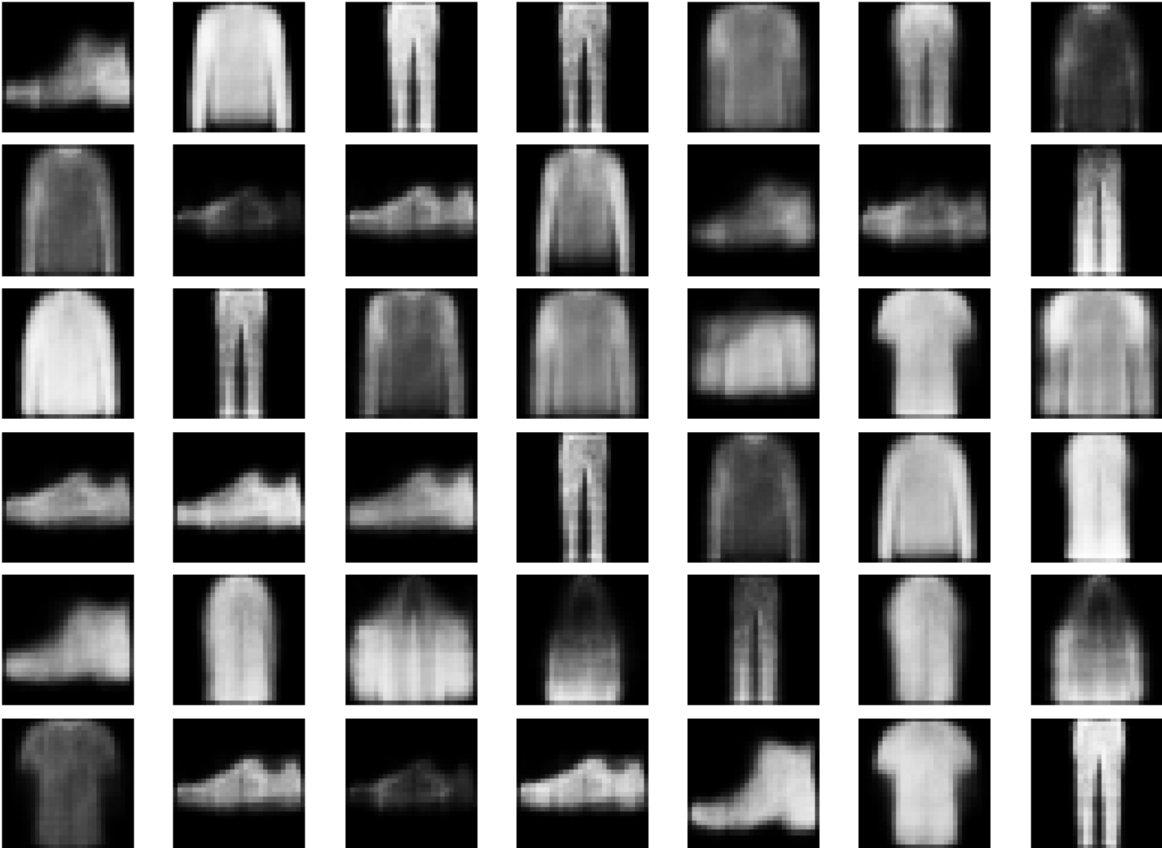}
        \caption{}
        \label{fig:main-a}
    \end{subfigure}
    \hfill
    \begin{subfigure}[b]{0.48\textwidth}
        \centering
        \includegraphics[width=\linewidth]{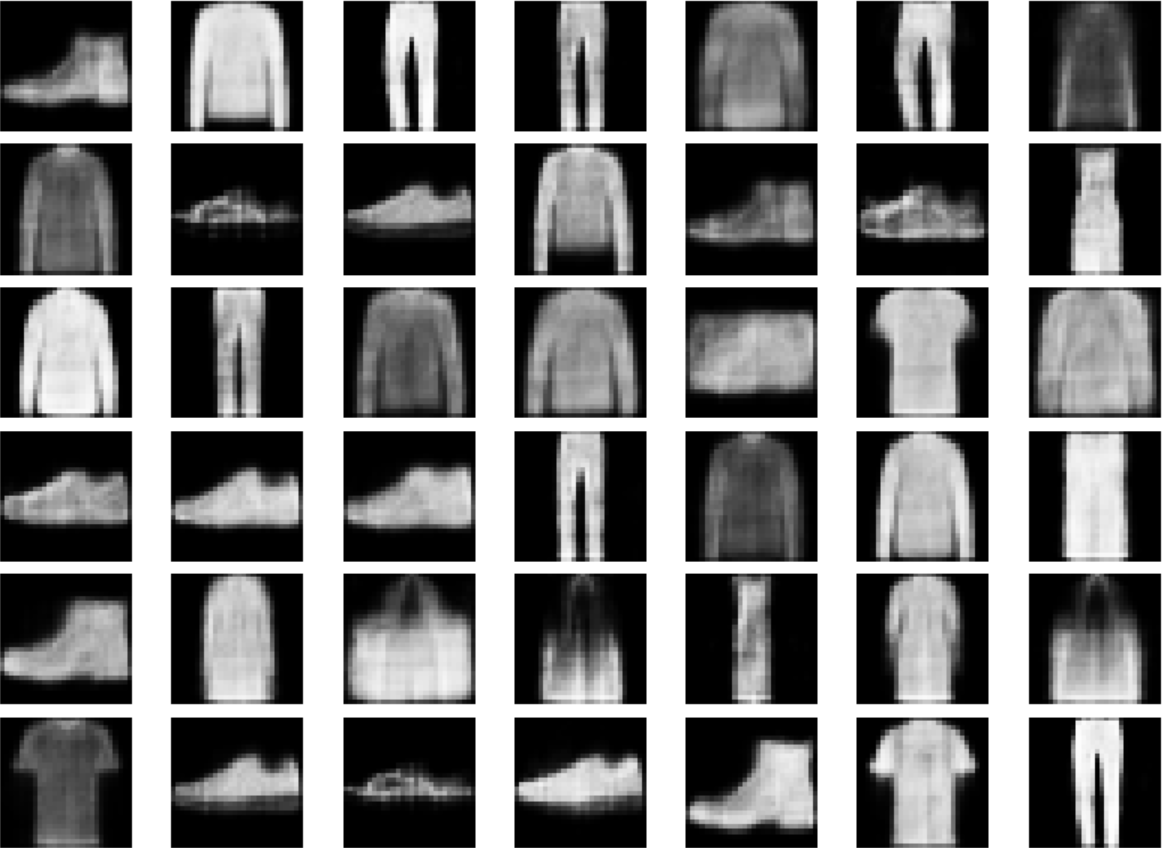}
        \caption{}
        \label{fig:main-b}
    \end{subfigure}
    \caption{Visual outputs of different readouts for the Fashion MNIST dataset without class-wise training. The results were obtained in (a) with the $\langle Z_i \rangle$ readout and in (b) with $\langle X_i \rangle$, $\langle Y_i \rangle$, $\langle Z_i \rangle$, $\langle Z_i Z_{i+1} \rangle$ readouts.}
    \label{fig:main}
\end{figure}

In Appendix B,  we use a different training model to reduce the tendency of the images to converge to the mean. Instead of class-wise training for the QINR-AE, we train all classes simultaneously, with each class trained on a 500-sample training set. 
Since this approach makes training slightly more difficult, we make some modifications in our calculations. 
We find that adding a learnable global scale $s$ defined by
\begin{equation}
\tilde{\mathbf{h}}_i =s\, \mathbf{h}_i,  \qquad s=e^{\rho}>0,
\label{eq:s}
\end{equation}
where $\rho$ is a learnable parameter (on a log scale) of the qubit rotation angle before the quantum circuit, can improve optimization. This parameter determines the total embedding magnitude and avoids vanishingly small rotation angles, leading to better gradient flow. Using $L=3$, $K=2$, and $epoch=40$, we obtain the results shown in Fig.~\ref{fig:main-a} for the Fashion MNIST dataset.

Furthermore, we add multiple readouts to enhance the features, for example, by adding multiple bases (multiobservable) $\langle X_i \rangle$, $\langle Y_i \rangle$, $ \langle Z_i \rangle$, and nearest neighbor correlator $\langle Z_i Z_{i+1} \rangle$ readouts. The image quality improves with this procedure, as shown in Fig.~\ref{fig:main-b}. A single observable provides only one projection of the final quantum feature state, whereas the combined  $\langle X_i \rangle$, $\langle Y_i \rangle$, $ \langle Z_i \rangle$ and $\langle Z_i Z_{i+1} \rangle$ readouts provide more information. This is because the system is sensitive to the phase information of the state and receives a correlation signal between the qubits. In simulations, the expectation values are determined from the final state without any intermediate wavefunction collapse. However, several circuit executions with different hardware-based measurements are needed.
To compare the readout effects quantitatively, we use the PSNR, SSIM, and cosine similarity metrics. 

\FloatBarrier
In Tables~\ref{tab:1m} and~\ref{tab:4m}, we see that there is an improvement in the model as the measurement is modified from $\langle Z_i \rangle$ to $\langle X_i \rangle$, $\langle Y_i \rangle$, $ \langle Z_i \rangle$, $\langle Z_i Z_{i+1} \rangle$ by looking at outputs of the two different models.
The outputs of all of the metrics obtained via $\langle X_i \rangle$, $\langle Y_i \rangle$, $ \langle Z_i \rangle$, and $\langle Z_i Z_{i+1} \rangle$ readouts in Table~\ref{tab:4m} are higher than the $\langle Z_i \rangle$ only readout shown in Table~\ref{tab:4m}. 
This is because multiple readouts capture further details, which are missed when only one readout is used.
\begin{table}[h!]
\centering
\small
\setlength{\tabcolsep}{3pt}
\renewcommand{\arraystretch}{1.15}
\resizebox{0.95\linewidth}{!}{
\begin{tabular}{l *{10}{c}}
\toprule
\textbf{Labels} &
{\textbf{T-shirt}} & {\textbf{Trouser}} & {\textbf{Pullover}} & {\textbf{Dress}} & {\textbf{Coat}} &
{\textbf{Sandal}} & {\textbf{Shirt}} & {\textbf{Sneaker}} & {\textbf{Bag}} & {\textbf{A.Boot}} \\
\midrule
PSNR  & 16.22 & 16.62 & 16.04 & 15.79 & 16.37 & 14.72 & 16.17 & 17.45 & 14.79 & 15.68 \\
SSIM  & 0.581 & 0.782 & 0.608 & 0.623 & 0.620 & 0.513 & 0.558 & 0.739 & 0.480 & 0.648 \\
CosSim  & 0.928 & 0.926 & 0.934 & 0.909 & 0.947 & 0.751 & 0.921 & 0.909 & 0.918 & 0.934 \\
\bottomrule
\end{tabular}
}
\caption{Results from various metrics obtained for the MNIST dataset with $ \langle Z_i \rangle$ readout.}
\label{tab:1m}
\end{table}
\begin{table}[h!]
\centering
\small
\setlength{\tabcolsep}{3pt}
\renewcommand{\arraystretch}{1.15}
\resizebox{0.95\linewidth}{!}{
\begin{tabular}{l *{10}{c}}
\toprule
\textbf{Labels} &
{\textbf{T-shirt}} & {\textbf{Trouser}} & {\textbf{Pullover}} & {\textbf{Dress}} & {\textbf{Coat}} &
{\textbf{Sandal}} & {\textbf{Shirt}} & {\textbf{Sneaker}} & {\textbf{Bag}} & {\textbf{A.Boot}} \\
\midrule
PSNR  & 17.09 & 18.75 & 16.67 & 16.98 & 17.03 & 15.24 & 16.86 & 18.08 & 15.07 & 16.44 \\
SSIM  & 0.652 & 0.863 & 0.640 & 0.714 & 0.655 & 0.608 & 0.609 & 0.776 & 0.521 & 0.701 \\
CosSim  & 0.939 & 0.957 & 0.941 & 0.931 & 0.954 & 0.797 & 0.932 & 0.921 & 0.924 & 0.946 \\
\bottomrule
\end{tabular}
}
\caption{Results from various metrics obtained for the MNIST dataset with $\langle X_i \rangle$, $\langle Y_i \rangle$, $ \langle Z_i \rangle$, and $\langle Z_i Z_{i+1} \rangle$ readouts.}
\label{tab:4m}
\end{table}

\FloatBarrier

\section{Controlled Ablation of the QINR Decoder against a Linear Decoder}\label{secA3}
\begin{figure}[!h]
\centering
\includegraphics[width=0.7\linewidth,height=0.7\textheight,keepaspectratio]{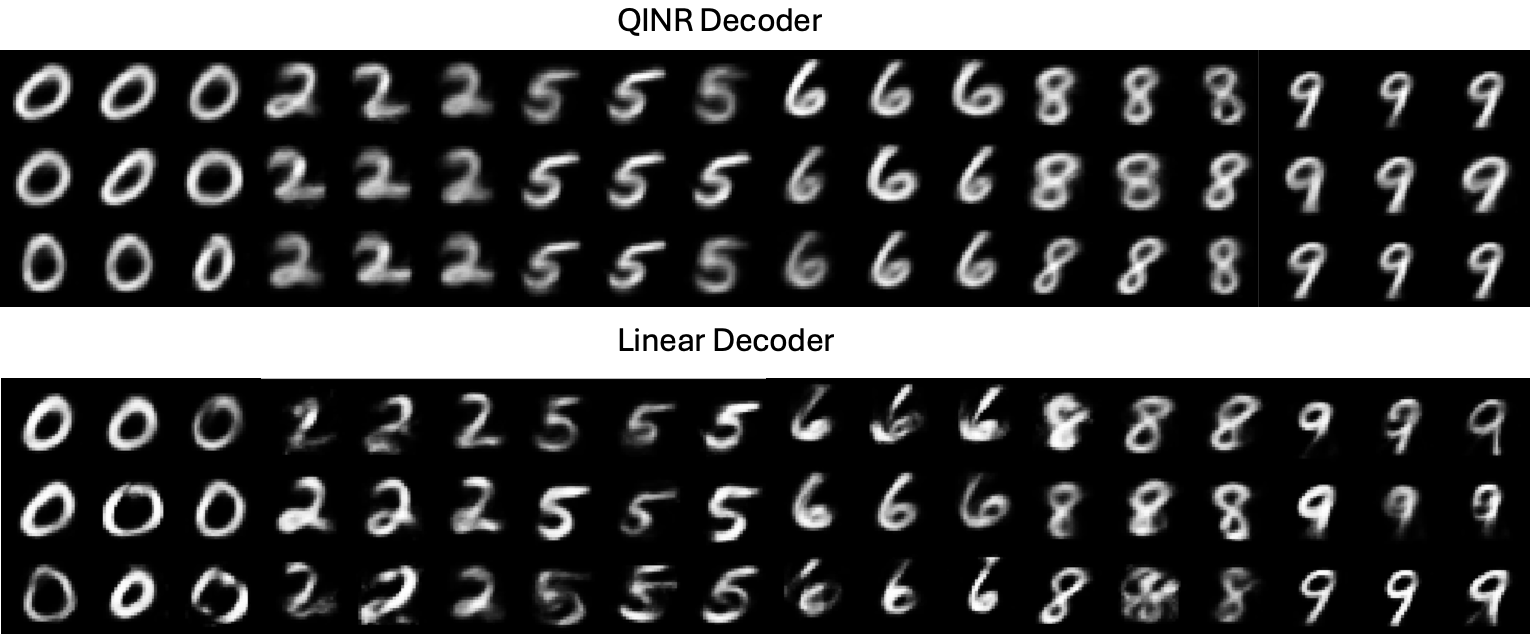}
\caption{\label{fig:lin}Images for the MNIST dataset generated by the QINR decoder and the classical decoder within VAE.}
\end{figure}

In Appendix C, to demonstrate the advantages of the QINR in the VAE, we also perform a controlled linear decoder ablation. This experiment is not intended as a comparison with optimized classical decoders, but as an ablation of the nonlinear spectral mapping introduced by the QINR component. The classical decoder we consider here consists of linear layers and has a similar number of parameters. We generate images for the MNIST dataset using both decoder models with $epochs=30$. The QINR decoder contains one BatchNorm, and the classical decoder contains two BatchNorms. In Fig.~\ref{fig:lin}, we show that the images generated by the classical decoder are discontinuous, whereas the images generated by a QINR decoder are more continuous. Although the classical model exhibits more variability, it is slightly more limited in the QINR model. 

We calculate the FID metrics for both the QINR and the linear decoder. We find that the results of the QINR are higher than those of the linear decoder models. The differences in the results per digit are shown in Table~\ref{tab:diff}. Considering that the diversity of the images is determined by the FID, these findings are reasonable. However, when we consider visual quality, we observe that the QINR decoder produces better results.
\FloatBarrier
\begin{table}[h!]
\centering
\small
\setlength{\tabcolsep}{3pt}
\renewcommand{\arraystretch}{1.15}
\resizebox{0.95\linewidth}{!}{
\begin{tabular}{l *{10}{c}}
\toprule
\textbf{Labels} &
{\textbf{0}} & {\textbf{1}} & {\textbf{2}} & {\textbf{3}} & {\textbf{4}} &
{\textbf{5}} & {\textbf{6}} & {\textbf{7}} & {\textbf{8}} & {\textbf{9}} \\
\midrule
QINR - LIN & 78.05 & 48.26 & 58.55 & 30.08 & 48.46 & 63.85 & 79.50 & 90.70 & 23.76 & 59.27 \\
\bottomrule
\end{tabular}
}
\caption{Difference in the FID metric between the QINR and the classical decoders.}
\label{tab:diff}
\end{table}
\FloatBarrier




\end{appendices}


\bibliography{reference}

\begin{thebibliography}{10}
\providecommand{\doi}[1]{\url{https://doi.org/#1}}
\bibcommenthead

\bibitem[\protect\citeauthoryear{Michelucci}{2022}]{michelucci2022introduction}
Michelucci U.
\newblock An introduction to autoencoders.
\newblock arXiv preprint arXiv:220103898. 2022;.

\bibitem[\protect\citeauthoryear{Vahdat and Kautz}{2020}]{NEURIPS2020_e3b21256}
Vahdat A, Kautz J.
\newblock NVAE: A Deep Hierarchical Variational Autoencoder.
\newblock In: Larochelle H, Ranzato M, Hadsell R, Balcan MF, Lin H, editors.
  Advances in Neural Information Processing Systems. vol.~33. Curran
  Associates, Inc.; 2020. p. 19667--19679.
\newblock Available from:
  \url{https://proceedings.neurips.cc/paper_files/paper/2020/file/e3b21256183cf7c2c7a66be163579d37-Paper.pdf}.

\bibitem[\protect\citeauthoryear{Goodfellow
  et~al.}{2014}]{goodfellow2014generative}
Goodfellow IJ, Pouget-Abadie J, Mirza M, Xu B, Warde-Farley D, Ozair S, et~al.
\newblock Generative adversarial nets.
\newblock Advances in neural information processing systems. 2014;27.

\bibitem[\protect\citeauthoryear{Croitoru et~al.}{2023}]{10081412}
Croitoru FA, Hondru V, Ionescu RT, Shah M.
\newblock Diffusion Models in Vision: A Survey.
\newblock IEEE Transactions on Pattern Analysis and Machine Intelligence.
  2023;45(9):10850--10869.
\newblock \doi{10.1109/TPAMI.2023.3261988}.

\bibitem[\protect\citeauthoryear{Higgins et~al.}{2017}]{higgins2017beta}
Higgins I, Matthey L, Pal A, Burgess C, Glorot X, Botvinick M, et~al.
\newblock beta-vae: Learning basic visual concepts with a constrained
  variational framework.
\newblock In: International conference on learning representations; 2017. .

\bibitem[\protect\citeauthoryear{Biamonte et~al.}{2017}]{biamonte2017quantum}
Biamonte J, Wittek P, Pancotti N, Rebentrost P, Wiebe N, Lloyd S.
\newblock Quantum machine learning.
\newblock Nature. 2017;549(7671):195--202.

\bibitem[\protect\citeauthoryear{Bravo-Prieto}{2021}]{bravo2021quantum}
Bravo-Prieto C.
\newblock Quantum autoencoders with enhanced data encoding.
\newblock Machine Learning: Science and Technology. 2021;2(3):035028.

\bibitem[\protect\citeauthoryear{Lloyd and
  Weedbrook}{2018}]{PhysRevLett.121.040502}
Lloyd S, Weedbrook C.
\newblock Quantum Generative Adversarial Learning.
\newblock Phys Rev Lett. 2018 Jul;121:040502.
\newblock \doi{10.1103/PhysRevLett.121.040502}.

\bibitem[\protect\citeauthoryear{Tsang et~al.}{2023}]{10264175}
Tsang SL, West MT, Erfani SM, Usman M.
\newblock Hybrid Quantum–Classical Generative Adversarial Network for
  High-Resolution Image Generation.
\newblock IEEE Transactions on Quantum Engineering. 2023;4:1--19.
\newblock \doi{10.1109/TQE.2023.3319319}.

\bibitem[\protect\citeauthoryear{Herr et~al.}{2021}]{Herr_2021}
Herr D, Obert B, Rosenkranz M.
\newblock Anomaly detection with variational quantum generative adversarial
  networks.
\newblock Quantum Science and Technology. 2021 jul;6(4):045004.
\newblock \doi{10.1088/2058-9565/ac0d4d}.

\bibitem[\protect\citeauthoryear{Dupont et~al.}{2021}]{dupont2021coin}
Dupont E, Goli{\'n}ski A, Alizadeh M, Teh YW, Doucet A.
\newblock Coin: Compression with implicit neural representations.
\newblock arXiv preprint arXiv:210303123. 2021;.

\bibitem[\protect\citeauthoryear{Zhao et~al.}{2024}]{pmlr-v235-zhao24}
Zhao J, Qiao W, Zhang P, Gao H.
\newblock Quantum Implicit Neural Representations.
\newblock In: Salakhutdinov R, Kolter Z, Heller K, Weller A, Oliver N, Scarlett
  J, et~al., editors. Proceedings of the 41st International Conference on
  Machine Learning. vol. 235 of Proceedings of Machine Learning Research. PMLR;
  2024. p. 60940--60956.
\newblock Available from:
  \url{https://proceedings.mlr.press/v235/zhao24l.html}.

\bibitem[\protect\citeauthoryear{Ma et~al.}{2025}]{ma2025}
Ma Q, Hao C, Si N, Chen G, Zhang J, Qu D.
\newblock Quantum adversarial generation of high-resolution images.
\newblock EPJ Quantum Technology. 2025;12(1):3.

\bibitem[\protect\citeauthoryear{Barsha and Eberle}{2025}]{barsha2025depth}
Barsha FL, Eberle W.
\newblock An in-depth review and analysis of mode collapse in generative
  adversarial networks.
\newblock Machine Learning. 2025;114(6):141.

\bibitem[\protect\citeauthoryear{Deng}{2012}]{6296535}
Deng L.
\newblock The MNIST Database of Handwritten Digit Images for Machine Learning
  Research [Best of the Web].
\newblock IEEE Signal Processing Magazine. 2012;29(6):141--142.
\newblock \doi{10.1109/MSP.2012.2211477}.

\bibitem[\protect\citeauthoryear{Cohen et~al.}{2017}]{7966217}
Cohen G, Afshar S, Tapson J, van Schaik A.
\newblock EMNIST: Extending MNIST to handwritten letters.
\newblock In: 2017 International Joint Conference on Neural Networks (IJCNN);
  2017. p. 2921--2926.

\bibitem[\protect\citeauthoryear{Xiao et~al.}{2017}]{xiao2017}
Xiao H, Rasul K, Vollgraf R.: Fashion-MNIST: a Novel Image Dataset for
  Benchmarking Machine Learning Algorithms.
\newblock Available from: \url{https://arxiv.org/abs/1708.07747}.

\bibitem[\protect\citeauthoryear{Liu et~al.}{2015}]{liu2015deep}
Liu Z, Luo P, Wang X, Tang X.
\newblock Deep learning face attributes in the wild.
\newblock In: Proceedings of the IEEE international conference on computer
  vision; 2015. p. 3730--3738.

\bibitem[\protect\citeauthoryear{Zhang et~al.}{2025}]{ZHANG2025107875}
Zhang J, Che X, Fan Y, Peng S, Chen G, Ma Q, et~al.
\newblock Denoising diffusion models with optimized quantum implicit neural
  networks for image generation.
\newblock Future Generation Computer Systems. 2025;173:107875.
\newblock \doi{https://doi.org/10.1016/j.future.2025.107875}.

\bibitem[\protect\citeauthoryear{Cordero et~al.}{2025}]{cordero2025quantum}
Cordero Y, Garc{\'\i}a-Molina P, Vilari{\~n}o F.
\newblock Quantum Implicit Neural Representations for 3D Scene Reconstruction
  and Novel View Synthesis.
\newblock arXiv preprint arXiv:251212683. 2025;.

\bibitem[\protect\citeauthoryear{Wang et~al.}{2026}]{wang2026quantum}
Wang S, Theobalt C, Golyanik V.
\newblock Quantum visual fields with neural amplitude encoding.
\newblock Advances in Neural Information Processing Systems.
  2026;38:158535--158559.

\bibitem[\protect\citeauthoryear{Arbib}{1998}]{lecun1998handbook}
Arbib M.: The handbook of brain theory and neural networks.
\newblock MIT press Cambridge, MA, USA:.

\bibitem[\protect\citeauthoryear{Schuld et~al.}{2021}]{PhysRevA.103.032430}
Schuld M, Sweke R, Meyer JJ.
\newblock Effect of data encoding on the expressive power of variational
  quantum-machine-learning models.
\newblock Phys Rev A. 2021 Mar;103:032430.
\newblock \doi{10.1103/PhysRevA.103.032430}.

\bibitem[\protect\citeauthoryear{Bergholm et~al.}{2022}]{bergholm2022}
Bergholm V, Izaac J, Schuld M, Gogolin C, Ahmed S, Ajith V, et~al.: PennyLane:
  Automatic differentiation of hybrid quantum-classical computations.
\newblock Available from: \url{https://arxiv.org/abs/1811.04968}.

\bibitem[\protect\citeauthoryear{Paszke et~al.}{2019}]{NEURIPS2019}
Paszke A, Gross S, Massa F, Lerer A, Bradbury J, Chanan G, et~al.
\newblock PyTorch: An Imperative Style, High-Performance Deep Learning Library.
\newblock In: Wallach H, Larochelle H, Beygelzimer A, d\textquotesingle
  Alch\'{e}-Buc F, Fox E, Garnett R, editors. Advances in Neural Information
  Processing Systems. vol.~32. Curran Associates, Inc.; 2019. Available from:
  \url{https://proceedings.neurips.cc/paper_files/paper/2019/file/bdbca288fee7f92f2bfa9f7012727740-Paper.pdf}.

\bibitem[\protect\citeauthoryear{Heusel et~al.}{2017}]{heusel2017gans}
Heusel M, Ramsauer H, Unterthiner T, Nessler B, Hochreiter S.
\newblock Gans trained by a two time-scale update rule converge to a local nash
  equilibrium.
\newblock Advances in neural information processing systems. 2017;30.

\bibitem[\protect\citeauthoryear{Wang et~al.}{2003}]{1292216}
Wang Z, Simoncelli EP, Bovik AC.
\newblock Multiscale structural similarity for image quality assessment.
\newblock In: The Thrity-Seventh Asilomar Conference on Signals, Systems \&
  Computers, 2003. vol.~2; 2003. p. 1398--1402 Vol.2.

\bibitem[\protect\citeauthoryear{Horé and Ziou}{2010}]{5596999}
Horé A, Ziou D.
\newblock Image Quality Metrics: PSNR vs. SSIM.
\newblock In: 2010 20th International Conference on Pattern Recognition; 2010.
  p. 2366--2369.

\bibitem[\protect\citeauthoryear{Zhu et~al.}{2018}]{zhu2018cosine}
Zhu X, Su S, Fu M, Liu J, Zhu L, Yang W, et~al.
\newblock A cosine similarity algorithm method for fast and accurate monitoring
  of dynamic droplet generation processes.
\newblock Scientific reports. 2018;8(1):9967.

\bibitem[\protect\citeauthoryear{Larocca et~al.}{2025}]{larocca2025barren}
Larocca M, Thanasilp S, Wang S, Sharma K, Biamonte J, Coles PJ, et~al.
\newblock Barren plateaus in variational quantum computing.
\newblock Nature Reviews Physics. 2025;7(4):174--189.

\end{thebibliography}

\end{document}